\documentclass{article}

\usepackage{iftex}

\ifPDFTeX
  \usepackage[T1]{fontenc}
  \usepackage[utf8]{inputenc}
  \usepackage{newtxtext}        
  \usepackage{newtxmath}      
\else
  \usepackage{fontspec}
  \usepackage{unicode-math}
    \usepackage[warnings-off={mathtools-colon,mathtools-overbracket}]{unicode-math}
\fi

\usepackage{amsmath}
\usepackage{mathtools}
\usepackage{xfrac}

\usepackage{microtype}
\usepackage{algorithm}
\usepackage{algorithmic}

\usepackage[hypertexnames=false, unicode, pdfusetitle,hidelinks]{hyperref}
\usepackage{xcolor}
\usepackage{graphicx}

\usepackage{booktabs}
\usepackage{siunitx}
\usepackage{etoolbox}
\robustify\bfseries

\usepackage{url}
\usepackage[capitalise, nameinlink, noabbrev]{cleveref}
\crefformat{equation}{(#2#1#3)}

\usepackage[utf8]{luainputenc}
\usepackage{csquotes}
\usepackage[english, main=english]{babel}
\usepackage{url}
\usepackage[capitalise, nameinlink, noabbrev]{cleveref}
\crefformat{equation}{(#2#1#3)}
\crefrangeformat{equation}{(#3#1#4) to~(#5#2#6)}
\crefmultiformat{equation}{(#2#1#3)}{ and~(#2#1#3)}{, (#2#1#3)}{ and~(#2#1#3)}

\usepackage[nonatbib, preprint]{neurips_2025}

\PassOptionsToPackage{numbers,sort&compress}{natbib}
\usepackage[numbers,sort&compress]{natbib}

\renewcommand{\d}{\,\mathrm{d}}

\DeclareMathOperator*{\argmin}{arg\,min}

\DeclarePairedDelimiterX{\Cond}[1]{(}{)}{
    
    #1
}

\makeatletter
\newcommand{\Fun}{\@ifstar\@sfun\@fun}
\newcommand{\@fun}[1]{#1\Cond}
\newcommand{\@sfun}[1]{#1\Cond*}
\makeatother

\DeclarePairedDelimiterX{\KLdelim}[2]{(}{)}{%
    #1\mkern2mu\delimsize\|\mkern2mu#2%
}

\DeclarePairedDelimiterXPP{\Moment}[2]{#1}{[}{]}{}{

#2
}

\DeclarePairedDelimiterX{\Set}[1]{\{}{\}}{
    
    #1
}

\DeclarePairedDelimiterXPP{\pix}[1]{\begingroup\scriptscriptstyle}{(}{)}{\endgroup}{\mkern-1mu#1\mkern-1mu}

\makeatletter
\DeclareRobustCommand{\rvdots}{%
    \vbox{
        \baselineskip4\p@\lineskiplimit\z@
        \kern-\p@
        \hbox{.}\hbox{.}\hbox{.}
    }}
\makeatother

\usepackage{listings}
\usepackage{wrapfig}
\usepackage{graphicx}
\usepackage{lipsum}
\usepackage{xcolor}
\usepackage[acronym]{glossaries}
\usepackage{makecell}

\usepackage{subcaption}
\usepackage{caption}
\usepackage{multirow}
\makeglossaries

\newacronym{MHD}{MHD}{magnetohydrodynamics}
\newacronym{QP}{QP}{quasi-poloidal}
\newacronym{QI}{QI}{quasi-isodynamicity}
\newacronym{GK}{GK}{gyro-kinetic}
\newacronym{ITG}{ITG}{ion-temperature gradient}
\newacronym{HV}{HV}{hypervolume}
\newacronym{LTS}{LTS}{low-temperature superconductors}
\newacronym{HTS}{HTS}{high-temperature superconductors}
\newacronym{QPS}{QPS}{quasi-poloidal stellarator}
\newacronym{W7X}{W7-X}{Wendelstein-7X}
\newacronym{ALM}{ALM}{Augmented Lagrangian method}
\newacronym{PCA}{PCA}{Principal Component Analysis}
\newacronym{MCMC}{MCMC}{Markov chain Monte Carlo}
\newacronym{GMM}{GMM}{Gaussian mixture model}

\newcolumntype{J}[1]{S[
  table-format=#1,
  round-mode=figures,
  round-precision = 4,
  scientific-notation = false,
  drop-zero-decimal=false
]}

\title{\texttt{ConStellaration}: A dataset of QI-like stellarator plasma boundaries and optimization benchmarks}

\author{
Santiago A. Cadena $\quad$
Andrea Merlo $\quad$
Emanuel Laude $\quad$
Alexander Bauer $\quad$
\\
\textbf{
Atul Agrawal $\quad$
Maria Pascu $\quad$
Marija Savtchouk $\quad$
Enrico Guiraud $\quad$
}
\\
\textbf{
Lukas Bonauer $\quad$
Stuart Hudson $\quad$
Markus Kaiser $\quad$
}
\\
\\
  Proxima Fusion \\
  \texttt{\{scadena, amerlo\}@proximafusion.com}
}

\begin{document}

\newcommand{\Nfp}{\ensuremath{N_{\mathrm{fp}}}}
\newcommand{\IotaOverNfp}{\ensuremath{\iota / \Nfp}}
\newcommand{\AveragedTriangularity}{\ensuremath{\bar{\delta}}}
\newcommand{\ie}{i.e.}
\newcommand{\eg}{e.g.}
\newcommand{\todo}[1]{\textcolor{red}{#1}}
\newcommand{\LGradB}{\ensuremath{L_{\nabla B}}}
\newcommand{\MirrorRatio}{\ensuremath{\Delta}}
\newcommand{\FluxCompression}{\ensuremath{\chi_{\nabla r}}}
\newcommand{\FluxSurfaceAveragedFluxCompression}{\ensuremath{\langle \FluxCompression \rangle}}
\newcommand{\mpol}{\ensuremath{m_{\text{pol}}}}
\newcommand{\ntor}{\ensuremath{n_{\text{tor}}}}
\newcommand{\params}{\ensuremath{\mathbf{\Theta}}}
\newcommand{\fig}[1]{Figure~\ref{#1}}
\newcommand{\vmec}{\texttt{VMEC}~}
\newcommand{\vmecpp}{\texttt{VMEC++}~}

\maketitle

\begin{abstract}

Stellarators are magnetic confinement devices under active development to deliver steady-state carbon-free fusion energy. 
Their design involves a high-dimensional, constrained optimization problem that requires expensive physics simulations and significant domain expertise. 
Recent advances in plasma physics and open-source tools have made stellarator optimization more accessible. 
However, broader community progress is currently bottlenecked by the lack of standardized optimization problems with strong baselines and datasets that enable data-driven approaches, 
particularly for quasi-isodynamic (QI) stellarator configurations, considered as a promising path to commercial fusion due to their inherent resilience to current-driven disruptions. 
Here, we release an open dataset of diverse QI-like stellarator plasma boundary shapes, 
paired with their ideal magnetohydrodynamic (MHD) equilibria and performance metrics. 
We generated this dataset by sampling a variety of QI fields and optimizing corresponding stellarator plasma boundaries.
We introduce three optimization benchmarks of increasing complexity: (1) a single-objective geometric optimization problem, (2) a ``simple-to-build" QI stellarator, and (3) a multi-objective ideal-MHD stable QI stellarator that investigates trade-offs between compactness and coil simplicity.
For every benchmark, we provide reference code, evaluation scripts, and strong baselines based on classical optimization techniques.
Finally, we show how learned models trained on our dataset can efficiently generate novel,
feasible configurations without querying expensive physics oracles.
By openly releasing the dataset (\url{https://huggingface.co/datasets/proxima-fusion/constellaration}) along with benchmark problems and baselines (\url{https://github.com/proximafusion/constellaration}), 
we aim to lower the entry barrier for optimization and machine learning researchers to engage in stellarator design
and to accelerate cross-disciplinary progress toward bringing fusion energy to the grid.

\end{abstract}

\section{Introduction}

\begin{wrapfigure}[8]{r}{0.25\textwidth}
  \centering
  \vspace*{-4em}
  \includegraphics[width=0.25\textwidth]{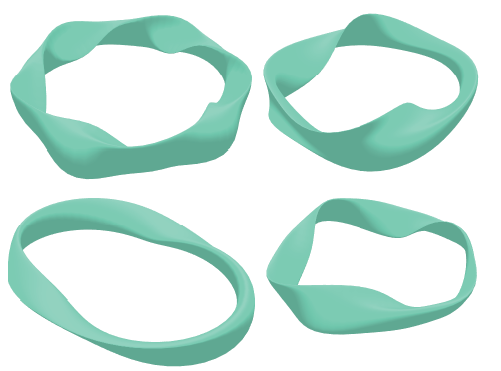}
  \caption{Examples of diverse stellarator plasma boundaries.}
  \label{fig:intro_examples}
\end{wrapfigure}

Fusion energy promises virtually limitless, carbon-free power by harnessing the same process that powers the sun. 
Magnetic confinement fusion approaches trap a fully ionized gas (plasma) within magnetic fields to sustain the conditions required for fusion.
Among these,
\emph{stellarators} confine the plasma solely through external coils,
which produce three-dimensional, twisted magnetic flux surfaces (\cref{fig:intro_examples}).
Unlike \emph{tokamaks},
stellarators do not rely on large internal plasma currents,
thereby avoiding associated instabilities~\cite{wesson2011tokamaks,hender2007mhd}.
However,
this advantage comes with a trade-off.
Designing stellarators involves a significantly more complex parameter space:
shaping the three-dimensional plasma boundary to satisfy multiple physics and engineering constraints is a high-dimensional, constrained,
optimization problem.

\begin{figure}[htbp]
  \centering
  \includegraphics[width=\textwidth]{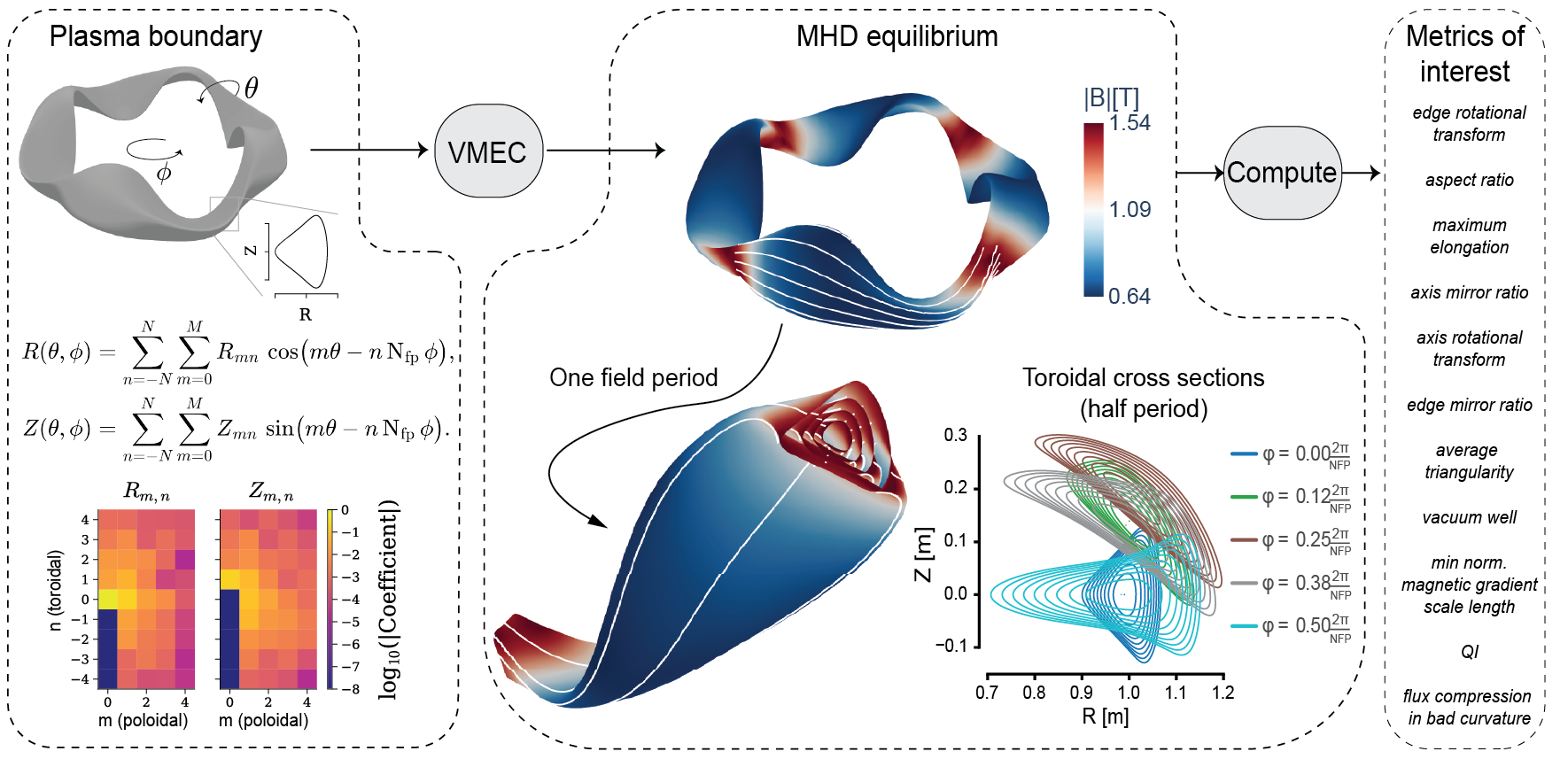}
  \caption{A plasma boundary is defined by the coefficients $R_{mn}$ and $Z_{mn}$ of a truncated Fourier series in cylindrical coordinates, parametrized by the lab-frame poloidal angle $\theta$ and toroidal angle $\phi$. This boundary is passed to the \vmecpp code \cite{hirshman1983steepest,schilling2025numerics} to compute an ideal-\gls{MHD} equilibrium. In this example, the configuration is stellarator symmetric, meaning that $R(\theta,\phi) = R(-\theta,-\phi)$ and $Z(\theta,\phi) = -\,Z(-\theta,-\phi)$, and the number of repeated field periods ($\Nfp$) is four.
  The ideal-\gls{MHD} equilibrium defines the magnetic field throughout the plasma volume,
  comprising nested magnetic flux surfaces on which magnetic field lines (depicted in white) lie.
  We can then compute various metrics of interest from the equilibrium field.}
  \label{fig:intro_mhd}
\end{figure}

Stellarator design has been mainly approached as a two-stage process~\cite{Henneberg_2021}.
In \textit{stage one},
the magnetic field that confines the plasma is optimized;
in \textit{stage two},
electromagnetic coils are designed to reproduce this field.
\textit{Stage one}, the focus of this work,
optimizes a three-dimensional surface that defines the boundary condition for the plasma equilibrium magnetic field.
The surface is commonly parameterized by a truncated Fourier series in cylindrical coordinates (\fig{fig:intro_mhd}, left).
A solution to the ideal-\gls{MHD} equations is then computed to determine the magnetic field inside the plasma~\cite{freidberg2014ideal}.
\vmec\cite{hirshman1983steepest} and its recent C++ re-implementation~\cite{schilling2025numerics} are classical physics codes that compute a solution to the ideal-\gls{MHD} model (Figure \ref{fig:intro_mhd}).
From the \gls{MHD} solution, we can compute multiple magnetic field properties,
\eg~the \textit{rotational transform},
that we can iteratively optimize to target a desired value in an outer optimization loop by updating the plasma boundary.

Unlike tokamaks,
classical or unoptimized stellarators lack toroidal symmetry and inherently suffer from poor confinement of high-energy particles:
such fusion-born particles often escape the plasma volume,
striking plasma facing components before depositing their energy back into the plasma.
This prevents a self-sustained fusion process.
The root cause lies in the behavior of particles trapped in poloidal, toroidal, or helical magnetic wells,
which fail to sample the entire magnetic flux surface,
experiencing a net radial drift that leads to gradually loosing confinement.
These challenges are a direct consequence of the non-axisymmetric magnetic geometry of stellarators.
A particularly effective strategy to suppress these drifts is to optimize stellarators imposing the condition of \textit{omnigeneity}~\cite{cary1997omnigenity,dudt2024magnetic},
which requires only that the average radial drift of trapped particles vanishes.
Among omnigenous fields,
\gls{QI} fields  have poloidally closed contours of the magnetic field strength~\cite{helander2009bootstrap,helander2014theory,goodman2023constructing},
which results in a vanishing net plasma toroidal current.
The advantages of even approximate \gls{QI} fields have been validated in laboratory experiments,
most notably in the \gls{W7X} stellarator~\cite{beidler2021demonstration}.
These compelling benefits have made the \gls{QI} symmetry a target in the design of next-generation stellarator-based fusion power plants \cite{lion2025stellaris,hegna2025infinity}.

Major advances in open-source software frameworks for stellarator design have been presented in recent years. 
For example, \texttt{SIMSOPT} \cite{landreman2021simsopt} provides high‐level interfaces to link plasma equilibrium solvers such as \vmec \cite{hirshman1983steepest} or \texttt{SPEC}
\cite{10.1063/1.4765691}
with numerical optimizers.
Moreover, tools like \texttt{DESC} \cite{dudt2020desc}
have leveraged end-to-end automatic differentiation \cite{blondel_elements_2024} to simultaneously compute MHD equilibria and target desired properties.
However, these tools still present a high entry barrier for practitioners in the optimization and machine learning communities,
as they require substantial domain knowledge to make meaningful contributions.

Although significant progress has been made in defining what to target in stellarator design, there remains a lack of standardized benchmark problems and evaluation protocols to address stage one optimization. This contrasts to other areas of machine learning, where well-defined challenges have driven rapid and measurable progress~\cite{hardt2025emerging}.
Establishing such benchmarks in stellarator research would offer significant value by enabling systematic comparisons of optimization methods across a range of problem formulations.
For instance, different representations (parameterizations) of the plasma boundary may vary in their effectiveness:
some may better avoid local minima,
while others may facilitate faster or more reliable convergence to feasible solutions.
Our contributions are as follows.

\begin{itemize}
    \item We release a diverse dataset of about 158,000 \gls{QI}-like stellarator plasma boundaries with their associated ideal-\gls{MHD} equilibria (in vacuum and for five different levels of plasma beta, the ratio between the plasma thermal pressure and the magnetic pressure) computed with \vmecpp~\cite{schilling2025numerics} and corresponding figures of merit.
    \item We propose three optimization problems of varying complexity and kind, and release associated code.
    \item We provide a set of baselines for these optimization problems using classical optimization approaches. 
    \item We show that models trained on our dataset can generate novel configurations that satisfy optimization constraints, even when only a handful of training examples do.
\end{itemize}

\subsection{Related work}

\paragraph{Stellarator datasets.}

Beyond works releasing a small set of plasma configurations \cite{nies2024exploration, buller2025family},
large-scale datasets have focused on stellarators with quasi-axisymmetry (QA) or quasi-helical symmetry (QH) \cite{landreman2022mapping, giuliani2024direct, giuliani2024comprehensive}, but not \gls{QI}.
These studies rely on an expansion about the magnetic axis \cite{mercier1964equilibrium, garren1991existence, garren1991magnetic, landreman2019optimized, landreman2019constructing, rodriguez2023constructing} (the field line representing the innermost flux surface) that reduces the 3D \GLS{MHD} equations into a 1D ordinary differential equation,
which is much faster to solve.
\citet{landreman2022mapping} sampled $\sim500k$ QA and QH configurations, while \citet{giuliani2024direct, giuliani2024comprehensive} sampled $\sim 370k$ QA and QH configurations as part of the \texttt{QUASR} dataset \footnote{\url{https://quasr.flatironinstitute.org}}.
To the best of our knowledge,
none of these datasets include publicly available computed ideal-\gls{MHD} equilibria.

\paragraph{Stellarator optimization benchmarks.}
While several studies have proposed sets of optimization problems to test optimization strategies or shape parameterizations (\eg \cite{henneberg2021representing}), 
and others have surveyed optimization approaches \cite{conlin2024stellarator},
there are no standardized benchmarks for stellarator optimization.

\subsection{Background}

\begin{figure}[htbp]
  \centering
  \includegraphics[width=0.8\textwidth]{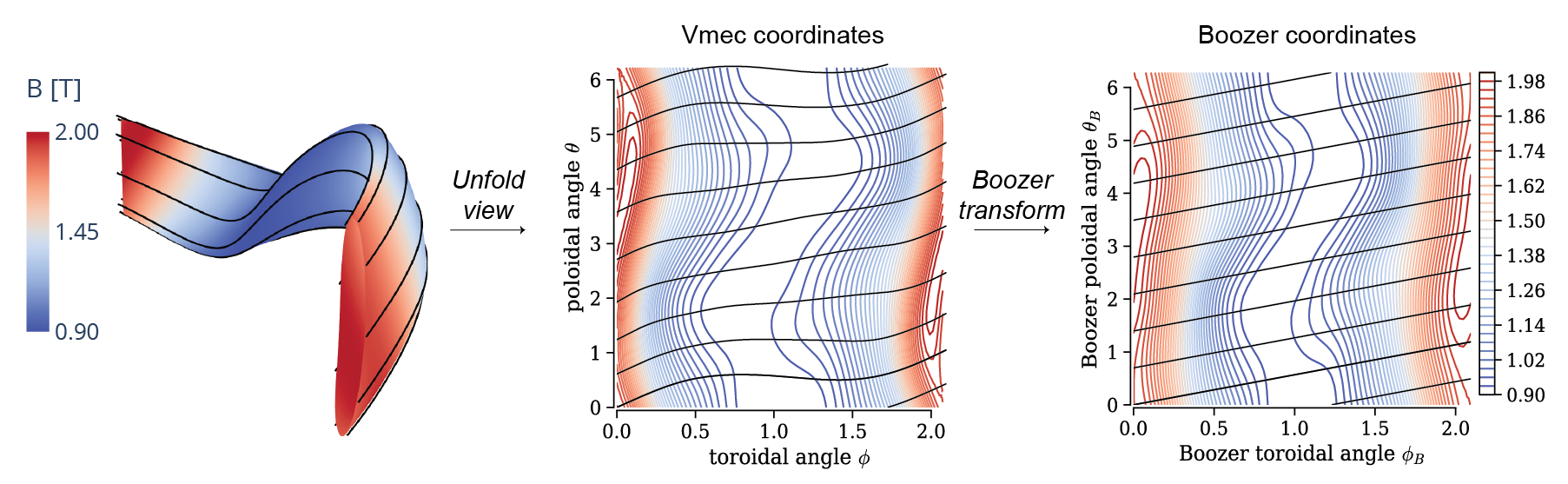}
  \caption{Visualization of the iso-contours of the magnetic field strength $B$ and a few magnetic field lines (black).
  In the \textit{Boozer} coordinate system~\cite{boozer1981plasma},
  the original poloidal and toroidal angles are transformed into Boozer angles $\theta_{B}$ and $\phi_{B}$, respectively, to straighten the magnetic field lines (black).}
  \label{fig:intro_mhd_boozer}
\end{figure}

In \textit{Boozer} coordinates~\cite{boozer1981plasma} (\cref{fig:intro_mhd_boozer}), 
magnetic field strength contours of QI fields exhibit three characteristic properties:
(i) the contours close poloidally, appearing as vertically closed loops in a Boozer plot;
(ii) the magnetic field strength maxima align along straight vertical lines; 
and (iii) the arc length between points of equal magnetic field strength along a field line depends only on the flux surface (\ie, it is invariant across field lines)~\cite{goodman2023constructing}.
The targets in \cref{fig:fig_dataset_examples} and \cref{fig:fig_diversity} are examples of precise QI fields.

\section{A diverse dataset of QI-like plasma boundaries and ideal-MHD equilibria}
\label{sec:data}

Directly sampling the Fourier coefficients representing the plasma boundary (\cref{fig:intro_mhd}) would very rarely lead to good (or even valid) stellarator fields~\cite{curvo2025using}. 
To generate a large and diverse dataset of stellarator configurations that are approximately \gls{QI}, 
we aim to sample diverse \gls{QI} fields and other geometrical properties,
and search for plasma boundaries that produce those target fields.
These target generative factors include the aspect ratio $A$ (the ratio between the major and minor toroidal radii: $R_0 / a$), 
the edge rotational transform $\iota_{\mathit{edge}}$ (how far a field line moves around the “short” (poloidal) way along the torus each time it goes once around the “long” (toroidal) way), 
the mirror ratio $\Delta_{\mathit{edge}}$ (defined as $(B_{\max} - B_{\min}) /(B_{\max} + B_{\min})$), 
and the maximum elongation $\epsilon_{\max}$ (the largest cross-section elongation across toroidal angles~\cite{goodman2023constructing}).

For a given target \gls{QI} field and set of properties, we generated surfaces either through physics-informed heuristics (Section 3 of~\citet{goodman2023constructing}), fast near-axis expansion models~\cite{landreman2019constructing, jorge2020near} (using \texttt{pyQSC} \footnote{\url{https://github.com/rogeriojorge/pyQIC}}), or through \textit{stage-one} optimization runs.
We passed all resulting surfaces to our forward model running \vmecpp at high fidelity (Section \ref{sec:benchmarks}) to obtain ideal-\gls{MHD} equilibria and metrics of interest (\cref{fig:intro_mhd}).
All configurations are limited to poloidal and toroidal mode numbers of at most four.
Assuming stellarator symmetry,
$R_{m, n} = 0; m=0, n< 0$ and $Z_{m, n} = 0; m=0, n \leq 0$,
and fixing the major radius $R_{0,0}=1$,
the total number of degrees of freedom is $80$ (\cref{fig:intro_mhd}, left).

\begin{figure}[htbp]
  \centering
  \includegraphics[width=\textwidth]{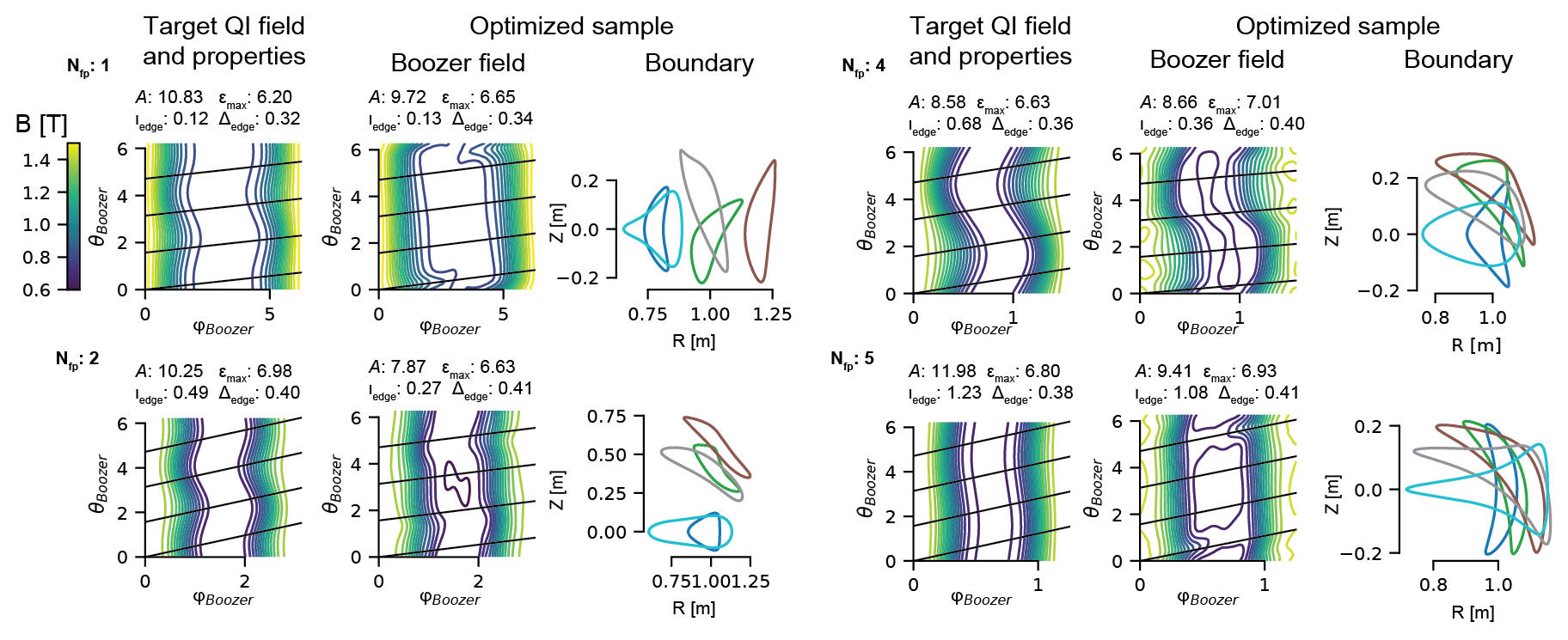}
  \vspace*{-2em}
  \caption{Four optimized samples from our dataset with 1, 2, 4, and 5 field periods. A finite computational budget for each sample generation leads to an approximate \gls{QI} field at the plasma boundary. All field plots share the same color bar and the boundary cross-section labels correspond to those in \fig{fig:intro_mhd}.}
  \label{fig:fig_dataset_examples}
\end{figure}

\begin{figure}[htbp]
  \centering
  \begin{minipage}[c]{0.75\textwidth}
    \includegraphics[width=\textwidth]{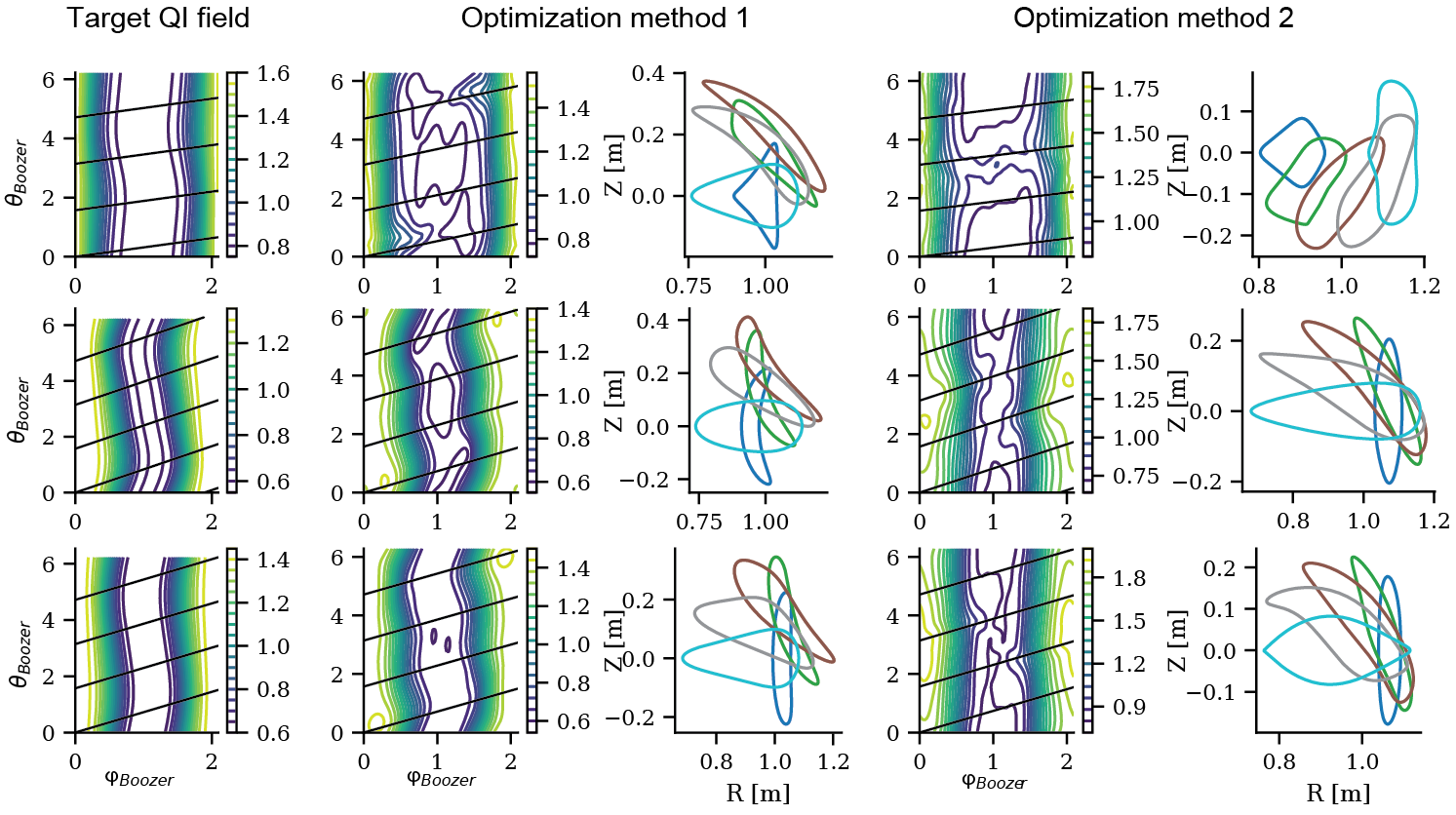}
  \end{minipage}\hfill
  \begin{minipage}[c]{0.22\textwidth}
    \caption{Diverse plasma configurations obtained for the same targets. Optimization methods vary in initialization strategy, framework, and settings. While some runs favor matching the target \gls{QI} field and mirror ratio, other runs better match the remaining target properties.}
    \label{fig:fig_diversity}
  \end{minipage}
\end{figure}

\begin{figure}[htbp]
  \centering
  \includegraphics[width=\textwidth]{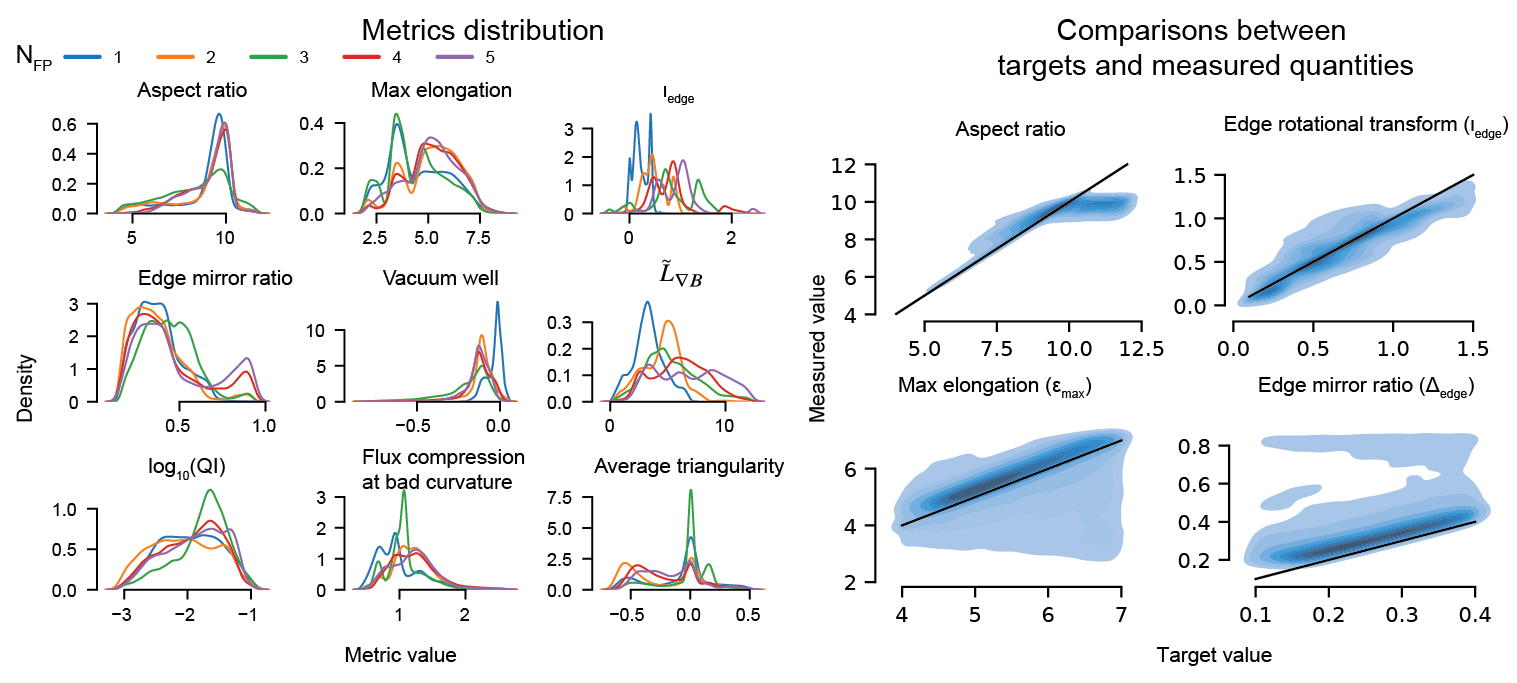}
  \vspace*{-2em}
  \caption{Distribution of metrics and comparisons between targets and outcomes.
  Pair plots show only optimized configurations. Black lines represent the identity. $A$, $\epsilon_{max}$, and $\Delta_{edge}$ were used as upper-bound constraints during optimization,
  while the rotational transform was enforced as equality constraint.}
  \label{fig:results}
\end{figure}

\paragraph{Sampling targets.}
To sample diverse QI fields,
we used the parameterization for an omnigenous field from \citet{dudt2024magnetic},
imposing stellarator symmetry (\cref{app:sampling}). 
Notably,
our fields span a diverse range of magnetic well shapes and show variation in how these wells are stretched along field lines  (\cref{fig:fig_dataset_examples}). 
The other target properties were drawn from a uniform distribution spanning a range of sensible values (\cref{tab:sampling_ramges}).
\paragraph{Optimization.} 
We implemented stage-one optimization approaches seeded with heuristic or near-axis expansion models using \texttt{DESC} or \vmecpp-based frameworks and varying objective settings (\cref{app:desc_optimization}).
Multiple optimization approaches with finite budget increased diversity in the resulting boundaries, 
even for the same set of target field and properties (\fig{fig:fig_diversity}). 
Each DESC run took three minutes on average on a 32 vCPU 128GB RAM machine,
while each \vmecpp run took around one hour on average on a 32 vCPU 32GB RAM machine.
\paragraph{Results.} 
We began by sampling 100k target sets. From this pool, we generated 30k and 49k plasma boundary candidates using our heuristic and near‐axis‐expansion models, respectively. We then applied the \texttt{DESC} optimizer twice to each target--once per initialization strategy--yielding an additional 88k optimized boundaries. A subset of 15k targets was also optimized with \vmecpp in the loop, seeded by rotating ellipses. Altogether, this produced roughly 182k candidate configurations,
and we evaluated equilibria and metrics with the high-fidelity forward model on 158k of them without errors.
Among these successful cases, 15k, 20k, 68k, 27k, and 28k configurations have 1, 2, 3, 4, and 5 field periods, respectively. Our resulting dataset spans a broad range of target metrics (Fig. \ref{fig:results}, left) and reveals strong correlations between prescribed targets and the achieved values (\cref{fig:results}, right). To enable investigations of equilibrium properties at finite pressure profiles (\ie, beyond vacuum), we also made available the ideal-MHD equilibria of the boundaries at five different volume-averaged $\beta$ values \footnote{$\beta$ is defined as the ratio of the thermal plasma pressure $p$ to the magnetic field pressure that has to be externally applied: $\beta=2\mu_0p/B^2$ where $\mu_0$ is the vacuum permeability. To set these beta values, we assumed a radial linear pressure profile and scaled the pressure at the axis to match the target volume-averaged beta for each boundary.} ($1,2,3,4,$ and $5\%$) and their correspondent metrics.
%
%

\paragraph{Statistical consistency of the dataset.}
To assess the degree to which the target metrics can be inferred from the available boundary coefficients, we trained an ensemble of multilayer perceptrons (MLP) model on the optimized dataset samples and evaluated its performance on a held-out test set. 
The model achieved good predictive accuracy ($R^2>0.97$ and RMSE<0.1 for all metrics; \cref{app:predictability_dataset}), indicating that the metrics are fairly learnable in-domain with an expressive enough model.
%

\section{Optimization benchmark}
\label{sec:benchmarks}

Stellarator design can be naturally formulated as a multi-objective constrained optimization problem~\cite{conlin2024stellarator}.
The objectives and constraints arise from both engineering and economic considerations (\eg, limiting the aspect ratio to achieve a compact device) as well as physics-based requirements (\eg, ensuring a stable \gls{MHD} plasma).
The design process involves translating stakeholder expectations into a consistent set of feasible requirements,
and navigating the trade-offs among conflicting objectives in a manner that aligns with the overarching design goals.

We introduce three prototypical stellarator design tasks with increasing complexity each involving different subsets of design metrics (\cref{tab:metrics}):
(1) \emph{Geometric},
(2) \emph{Simple-to-Build \gls{QI}},
and (3) \emph{\gls{MHD}-stable \gls{QI}}, detailed in Table \ref{tab:optimization_problems}.

\paragraph{Forward Model}

We leverage \vmecpp\cite{schilling2025numerics} to compute vacuum \num{3}D ideal-\gls{MHD} equilibria,
scaled to $R_{0}=\SI{1}{\meter}$,
$B_{0} \simeq \SI{1}{\tesla}$.
Each vacuum equilibrium is fully defined by a single flux‐surface mapping $\Sigma_{\Theta} : (\theta,\varphi) \; \longmapsto \; (R,\phi,Z)$,
where $\theta$ and $\varphi$ are generic poloidal and toroidal angles,
respectively,
and $\Theta$ denotes the set of surface parameters,
and $(R,\phi,Z)$ are cylindrical coordinates.
In \vmecpp,
the truncated Fourier series presented in~\cref{fig:intro_mhd} is used for $\Sigma_{\Theta}$.
However,
for the purpose of these optimization benchmark problems,
we make no assumptions about the functional form of $\Sigma_{\Theta}$.
All optimization problems have the form:

\begin{equation}
\begin{aligned}
    & \min_{\Theta} \; \left(f_1( \Theta ), f_2(\Theta), \ldots\right) \\
    & \text{subject to} \quad 
    c_i(\Theta) \leq c_i^{\ast}, \; \forall i \; ,
\end{aligned}
\label{eq:multiobjective_opt}
\end{equation}
where $f_i: \mathbb{R}^D \rightarrow \mathbb{R}$ are objective functions,
$c_i: \mathbb{R}^D \rightarrow \mathbb{R}$ are constraint functions,
and $c_i^{\ast}$ are constraint targets.
Each objective and constraint depends directly on the magnetic field,
which in turn is determined by the surface mapping that defines the boundary condition of the ideal-\gls{MHD} model.

\begin{wraptable}[10]{r}{0.65\textwidth}  
  \centering
  \footnotesize
  \vspace{-3em}
\begin{tabular}{c|c}
\hline
    Metric & Acronym \\
\hline
 minimum normalized magnetic gradient scale length  & $\widetilde{L}_{\nabla \mathbf{B}}$\\
  edge rotational transform over number of field periods & $\tilde\iota$ \\
  aspect ratio &  $A$ \\
  max elongation & $\epsilon_{\max}$ \\
  edge magnetic mirror ratio  & $\Delta$ \\
  quasi isodynamicity residual & $\mathit{QI}$ \\
  vacuum well & $W_{\gls{MHD}}$ \\
  flux compression in regions of bad curvature & $\FluxSurfaceAveragedFluxCompression$ \\
  average triangularity & $\bar \delta$\\
  \hline
\end{tabular}
\vspace{0.5em}
\caption{\label{tab:metrics} Equilibrium field metrics and their acronyms.}
\end{wraptable}

\begin{table}[htbp]
  \normalsize
  \begin{subtable}[t]{0.32\textwidth}
    \caption{\textbf{Geometric problem}}
    \vspace{-1.8em}
    \begin{alignat*}{2}
      &\min_{\Theta} \; && \epsilon_{\max} \\
      &\text{s.t.} \;       && A \le A^{*}, \\
      &&& \bar{\delta} \le \bar{\delta}^{*}, \\
      &&& \tilde{\iota} \ge \tilde{\iota}^{*}.
    \end{alignat*}
  \end{subtable}
  \hfill
  \begin{subtable}[t]{0.32\textwidth}
    \caption{\textbf{Simple-to-build QI}}
    \vspace{-1.8em}
    \begin{alignat*}{2}
      &\min_{\Theta} \; && -\widetilde{L}_{\nabla B} \\
      &\text{s.t.} \; && \tilde{\iota} \ge \tilde{\iota}^{*}, ~~\mathit{QI} \le \mathit{QI}^{*} \\
      &&& \Delta \le \Delta^{*}, ~~ A \le A^{*} \\
      &&& \epsilon_{\max} \le \epsilon_{\max}^{*}
    \end{alignat*}
  \end{subtable}
  \hfill
  \begin{subtable}[t]{0.32\textwidth}
    \caption{\textbf{MHD-stable QI}}
    \vspace{-1.8em}
    \begin{alignat*}{2}
  &\min_{\Theta} \; && \bigl(-\widetilde{L}_{\nabla B},\; A\bigr) \\
  &\text{s.t.} \; && \tilde{\iota} \ge \tilde{\iota}^{*},~~ \mathit{QI} \le \mathit{QI}^{*} \\
  &&& \Delta \le \Delta^{*}, ~~ W_{\mathrm{MHD}} \ge 0 \\
  &&& \langle \chi_{\nabla r} \rangle \le \langle \chi_{\nabla r} \rangle^{*}
\end{alignat*}
  \end{subtable}
  \caption{Constrained optimization problem formulations. See \cref{tab:metrics} for semantic associations to the symbols. All metrics are a function of the boundary}
\label{tab:optimization_problems}
\end{table}

\vspace{-2em}
\subsection{Problem 1: Geometric}

To onboard contributors to stellarator optimization, we propose an intuitive, purely geometric problem (\cref{tab:optimization_problems}) where
we look for stellarators that minimize the maximum elongation $\epsilon_{max}$ for a given aspect ratio $A$,
edge rotational transform $\tilde{\iota}$,
and average triangularity $\AveragedTriangularity$.
$\AveragedTriangularity$ averages the triangularity between the two stellarator-symmetric cross-sections
($\phi=0$ and $\phi=\pi / \Nfp$),
and $\tilde \iota$ is the edge rotational transform per field period.

\subsection{Problem 2: Single‐objective simple-to-build \gls{QI} stellarator}

Stellarators are notoriously challenging to construct due to their inherently three-dimensional magnetic geometry.
Optimized designs like \gls{W7X} demand millimeter coil tolerances~\cite{rummel2004accuracy}.
Moreover, the development and assembly of such devices can run into cost and schedule overruns driven by manufacturing complexity, potentially leading to the cancellation of entire projects as it was the case for the NCSX stellarator~\cite{NCSXCloseout2009,neilson2010lessons}.
This raises a key question:
\textit{Can optimized \gls{QI} stellarators be realized using simpler,
easier-to-manufacture coils?}

In a fusion reactor,
the spatial region between the plasma and coils must accommodate a divertor,
first wall (plasma-facing material components),
neutron shielding,
tritium-breeding blanket,
and magnets structural support.
These layers,
together with the magnet superconducting technologies
(\eg, \gls{LTS} or \gls{HTS}),
impose geometric and engineering constraints on coil design.
The feasibility of a stellarator configuration depends not only on plasma performance but also on how easily the required magnetic fields can be generated using manufacturable coils.

Not all magnetic fields are equally \textit{coil-friendly}.
We colloquially refer to \emph{coil simplicity} as the ease with which modular coils can be placed and shaped to produce the desired field.
For example,
surfaces with high coil simplicity allow coils to be located further from the plasma and require lower curvature and fewer tight bends.
We quantify coil simplicity using the normalized magnetic field gradient scale length on the plasma boundary,
following~\citet{kappel2024magnetic}.
This metric has proven effective in guiding optimization towards configurations with simpler,
more feasible coil designs~\cite{lion2025stellaris,hegna2025infinity}.

Historically,
\gls{QI} stellarators have required particularly complex coil geometries compared to other quasi-symmetric configurations~\cite{liu2018magnetic,jorge2022single,wechsung2022precise,wiedman2023coil}. This benchmark problem challenges that assumption by optimizing for \emph{precise} \gls{QI} fields that can be generated with \emph{simple} coils.

\Cref{tab:optimization_problems} introduces the problem definition,
where $\widetilde{L}_{\nabla \mathbf{B}}$ is magnetic field gradient scale length~\cite{kappel2024magnetic} normalized by $a/N_{\mathrm{fp}}$,
$\mathit{QI} = \frac{1}{4 \pi^2} \int \int r_{QI}^2 \d \theta \d\phi$ quantifies deviation from a \textit{precise} \gls{QI} field following~\citet{goodman2023constructing},
and $\Delta$ is the magnetic mirror ratio at the plasma boundary.
We normalize the objective by $a/N_{\mathrm{fp}}$ to ensure scale invariance across configurations with varying field period numbers.
Since a \gls{QI} field is easier to achieve for large aspect ratio configurations,
highly elongated flux surfaces,
and large mirror ratios~\cite{goodman2023constructing},
we explicitly control these quantities through inequality constraints.

\subsection{Problem 3: Multi-objective ideal-\gls{MHD} stable \gls{QI} stellarators}

This optimization problem introduces two new critical constraints for reactor relevant stellarator design: ideal-\gls{MHD} plasma stability and mitigation of turbulent transport.

Despite the fact that \gls{QI} configurations eliminate current-driven instabilities (``disruptions") that often affect tokamak designs, pressure-driven instabilities persist~\cite{helander2009bootstrap}, thus limiting access to high fusion power density regimes. To optimize for ideal‐\gls{MHD} stability, we adopt the vacuum magnetic well $W_{\text{MHD}}$ as a proxy \cite{mercier1964equilibrium,greene1997brief}

Turbulent transport,
expected to be dominated by \gls{ITG} turbulence in \gls{QI} stellarators~\cite{beidler2021demonstration,beurskens2021ion,goodman2024quasi},
limits the achievable fusion gain.
\citet{landreman2025does} demonstrated how purely geometrical quantities correlate strongly with the turbulence heat flux.
As a constraint, we compute the ``flux‐surface compression in regions of \textit{bad curvature}" given by
$\chi_{\nabla r} = \mathcal{H} (\mathbf{B} \times \mathbb{\kappa} \cdot \nabla \alpha) \left \lVert \nabla r \right \rVert_2^2 $
as a simple geometric proxy.
Here $\mathcal{H}$ is the Heavyside step function,
$\mathbf{B} \times \mathbb{\kappa} \cdot \nabla \alpha$ is the curvature drift~\cite{roberg2024reduction}
, $\mathbb{\kappa}$ is the magnetic field curvature, $\alpha$ is the field line label,
and $\nabla r$ is the flux compression.
A positive curvature drift represents regions of \textit{bad curvature}.
This quantity is evaluated on a single-flux surface at $\rho=r/a=0.7$.

In \gls{QP} and \gls{QI} stellarators,
$L_{\nabla \mathbf{B}} \propto R_0 / N_{\mathrm{fp}} = a A / N_{\mathrm{fp}}$
\footnote{
    Assuming that the characteristic length scale of the magnetic field gradient satisfies $L_{\nabla \mathbf{B}} \propto L_{\nabla B}$,
    and considering a \gls{QP} magnetic field where the magnetic field strength forms a single well,
    \ie,
    $B(\varphi) = B_0 \cos(N_{\mathrm{fp}} \varphi)$,
    where $\varphi$ is a field-aligned coordinate.
}.
More compact devices (low $A$) reduce capital cost per unit output power~\cite{spears1980scaling,freidberg2015designing} but increase coil complexity (proxied by $L_{\nabla \mathbf{B}}$).
This trade‐off motivates a Pareto‐optimal search \cite{martins2021engineering} between coil simplicity and compactness. \Cref{tab:optimization_problems} introduces the problem definition,
where $\langle \cdot \rangle$ denotes flux‐surface averaging.

\subsection{Evaluation metric}

We release evaluation code that scores candidate plasma boundaries across benchmarks.
Our evaluation code requires the plasma boundaries to be represented by the truncated Fourier series in cylindrical coordinates (see~\cref{fig:intro_mhd}).

\paragraph{Single-objective scoring}  
For single-objective problems,
we map each design point to a bounded scalar score value $s(\Theta)$ given by:
\begin{equation}\label{eq:single-objective-score}
  s(\Theta) = \begin{cases} h\bigl(f(\Theta)\bigr) & \text{if $\tilde{c}_i(\Theta) \leq \varepsilon,\; \forall i$}, \\
  0 & \text{otherwise,}
  \end{cases}
\end{equation}
where $f(\Theta)$ is the objective value,
$h:\mathbb{R}\rightarrow[0,1]$ is a linear map that rescales objectives into the $\left[ 0, 1 \right]$ interval (higher is better),
$\tilde{c}_i=(c_i-c_i^\ast)/c_i^\ast$ is the $i\text{-th}$ normalized constraint violation,
and $\varepsilon$ is a relative tolerance.

\paragraph{Multi-objective scoring}

For multi-objective problems,
we compute the \gls{HV} indicator~\cite{zitzler1998multiobjective,li2019quality} over feasible solutions
(\ie,
those with $\tilde{c}_i(\Theta) \leq \varepsilon,\; \forall i$
)
using a fixed reference point in objectives space.

\section{Optimization baselines}
\label{sec:baselines}
We now provide baselines for the three optimization problems.
For all experiments, we target stellarators with three field periods and seed optimizations from rotating ellipse configurations.
For the single-objective case (problem 1 and 2),
we benchmark three approaches:
a) gradient-based (where the gradient of the objective and constraint functions is approximated via forward finite-differences) 
trust-region interior point constrained optimizer~\cite{byrd1999interior} (\texttt{scipy-trust-constr});
b) gradient-free \texttt{COBYQA}~\cite{ragonneau2022model} algorithm (\texttt{scipy-COBYQA});
and c) \gls{ALM}~\cite{hestenes1969multiplier,powell1969method} with a non-Euclidean proximal regularization~\cite{rockafellar1976augmented,laude2023anisotropic, dufosse2021augmented} employing the \texttt{NGOpt} gradient-free meta-algorithm from Nevergrad~\cite{nevergrad} (\texttt{ALM-NGOpt}), to solve the subproblem.
Implementation specifics are provided in the \cref{app:optimization_baselines}.

\begin{wraptable}[7]{r}{0.35\textwidth}  
  \centering
  \footnotesize
  \vspace{-1em}
  \begin{tabular}{@{}lc@{}}
    \toprule
    Problem            & Score $\uparrow$ \\
    \midrule
    Geometrical        & 0.969 \\
    Simple-to-build    & 0.431 \\
    MHD-stable         & 130.0 \\
    \bottomrule
  \end{tabular}
  \vspace{0.1em}
  \caption{\texttt{ALM-NGOpt} scores.}
  \label{tab:scores}
\end{wraptable}

\begin{wrapfigure}[11]{r}{0.6\textwidth}
  \centering
  \vspace*{-3em}
  \begin{minipage}[b]{0.25\textwidth}
    \centering
    \includegraphics[width=\linewidth]{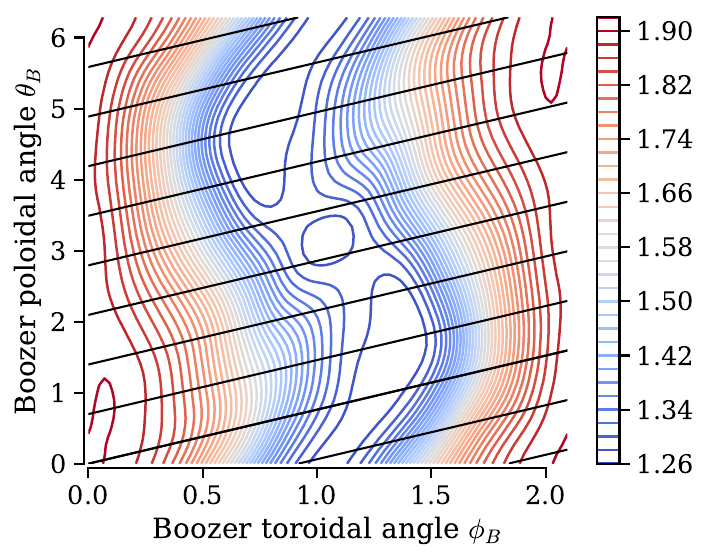}
  \end{minipage}%
  \hfill
  \begin{minipage}[b]{0.33\textwidth}
    \centering
    \includegraphics[clip, trim=1cm 4cm 0cm 4cm, width=\linewidth]{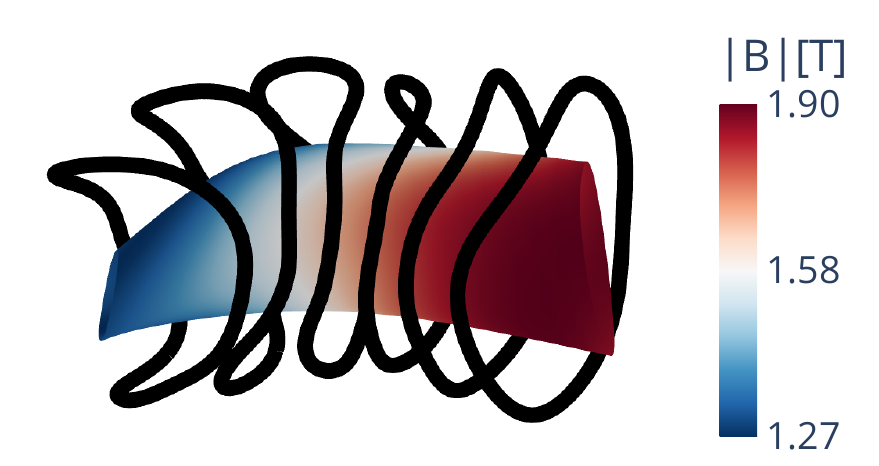}
  \end{minipage}
  \caption{
    Left: Optimized \gls{QI} magnetic field contours at the plasma boundary in Boozer coordinates for the simple-to-build optimization problem.
    Right: A representative coilset designed to reproduce the target magnetic field.
  }
  \label{fig:simple-to-build-boozer-and-coils}
\end{wrapfigure}

Only \texttt{ALM-NGOpt} obtains feasible solutions,
while both \texttt{scipy-trust-constr} and \texttt{scipy-COBYQA} did not (Table~\ref{tab:comparison_optimization_methods} and \ref{tab:scores}).
Consequently,
our leaderboard (Table~\ref{tab:scores}) reports results exclusively for \texttt{ALM-NGOpt}.
~\Cref{fig:simple-to-build-boozer-and-coils} shows the optimized \gls{QI} field and a representative coilset for the simple-to-build problem. 

\begin{figure}[htbp]
  \centering
  \begin{minipage}[c]{0.70\textwidth}
    \includegraphics[width=\linewidth]{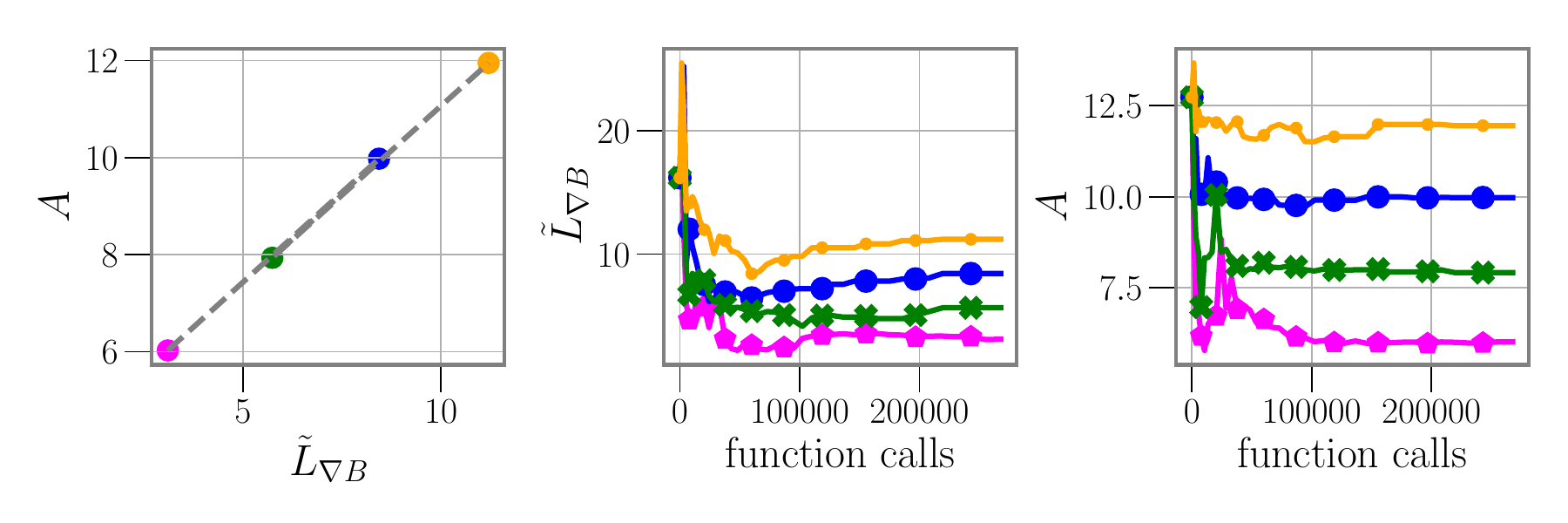}
  \end{minipage}\hfill
  \begin{minipage}[c]{0.28\textwidth}
    \caption{Pareto-front for the multi-objective optimization problem of \gls{MHD} stable QI stellarators obtained with \texttt{ALM-NGOpt}.}
    \label{fig:pareto_front}
  \end{minipage}
\end{figure}

The multi-objective problem is decomposed into multiple single-objective problems by treating the aspect ratio as an inequality constraint.
Using \texttt{ALM-NGOpt},
we found solutions for four of these instances.
A sparse Pareto front is provided in~\cref{fig:pareto_front}.

\begin{table}[h]
\centering
\footnotesize
\begin{tabular}{l | cc | cc}
\toprule
\multirow{2}{*}{Method} & \multicolumn{2}{c|}{Simple-to-build} & \multicolumn{2}{c}{Geometric problem} \\
\cmidrule(r){2-3} \cmidrule(l){4-5}
& $\tilde L_{\nabla B}\uparrow$ & norm. constr. viol. & $\epsilon_{\max}\downarrow$ & norm. constr. viol. \\
\midrule
\texttt{scipy-trust-constr} & 2.10$^{\ast}$ & 3.25$^{\ast}$ & 15.0$^{\ast}$ & 0.301$^{\ast}$ \\
\texttt{scipy-COBYQA}       & 14.4$^{\ast}$ & 2.04$^{\ast}$ & 1.27 & 0.953 \\
\texttt{ALM-NGOpt}      & 8.61 & 0.009 & 1.27 & 0.0002 \\
\bottomrule
\end{tabular}
\vspace{0.5em}
\caption{\label{tab:comparison_optimization_methods}Comparison of baselines for the simple-to-build and the geometric problem. $\uparrow$ means that a quantity is maximized and $\downarrow$ means that a quantity is minimized.
Final optimized boundaries for which \vmecpp failed to converge at high fidelity (\ie, the fidelity with which we score a plamsa boundary) are represented with $^{\ast}$;
for them, we report the objective and constraint values from a lower fidelity equilibrium computation.
\texttt{scipy-trust-constr} and \texttt{scipy-COBYQA} do not produce feasible solutions.
SciPy-based optimizers ran for $\sim40$ hours on a machine with 4 vCPUs.
\texttt{ALM-NGOpt} ran on a 96 vCPU machine for 18 hours (geometric problem) and 34 hours (simple-to-build).
}
\end{table}
\vspace{-0em}

\begin{table}[htbp]
  \centering
  \footnotesize
  \begin{minipage}[t]{0.20\textwidth}
    \vspace{0pt}                  
    \centering
    \begin{tabular}{ccc}
      \toprule
      $A\downarrow$ & $\tilde L_{\nabla B}\uparrow$ & norm.\ constr.\ viol. \\
      \midrule
      6.02  &  2.98   & 0.104   \\
      7.93  &  5.60   & 0.00130 \\
      9.98  &  8.45   & 0.0     \\
      11.9  &  11.1   & 0.00210 \\
      \bottomrule
    \end{tabular}
  \end{minipage}%
  \hfill
  \begin{minipage}[t]{0.60\textwidth}
    \vspace{-1em}                  
    \captionof{table}{%
      Objectives and constraint violations for \texttt{ALM-NGOpt} on the multi-objective problem.
      Optimization was carried out by solving a sequence of single-objective problems,
      converting one objective into a constraint $A \le A^*$ with $A^*\!\in\{6,8,10,12\}$.
      All instances were run on a 96-vCPU machine for 15–24\,h.
    }
    \label{tab:multi_objective}
  \end{minipage}
\end{table}

\section{Generative modeling of feasible domains without access to the oracle}
\label{sec:gen_model}

We present a method to generate feasible configurations using learning-based models trained on the dataset, 
without relying on a zero-order oracle (\eg, \vmecpp) and with limited feasible examples. 
We test whether this method can produce many valid configurations to support downstream tasks like optimization.

\begin{figure}[!h]
  \centering
  \begin{subfigure}[t]{0.475\textwidth}
    \centering
    \includegraphics[width=\linewidth]{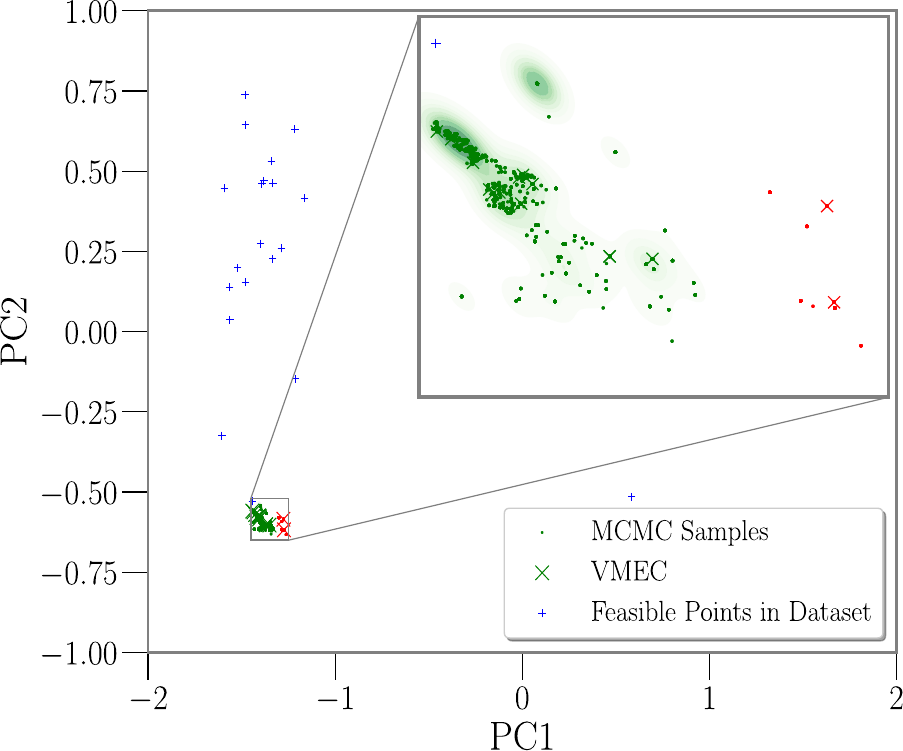}
    \caption{geometric problem.}
    \label{fig:posterior-a}
  \end{subfigure}\hfill
  \begin{subfigure}[t]{0.525\textwidth}
    \centering
    \includegraphics[width=\linewidth]{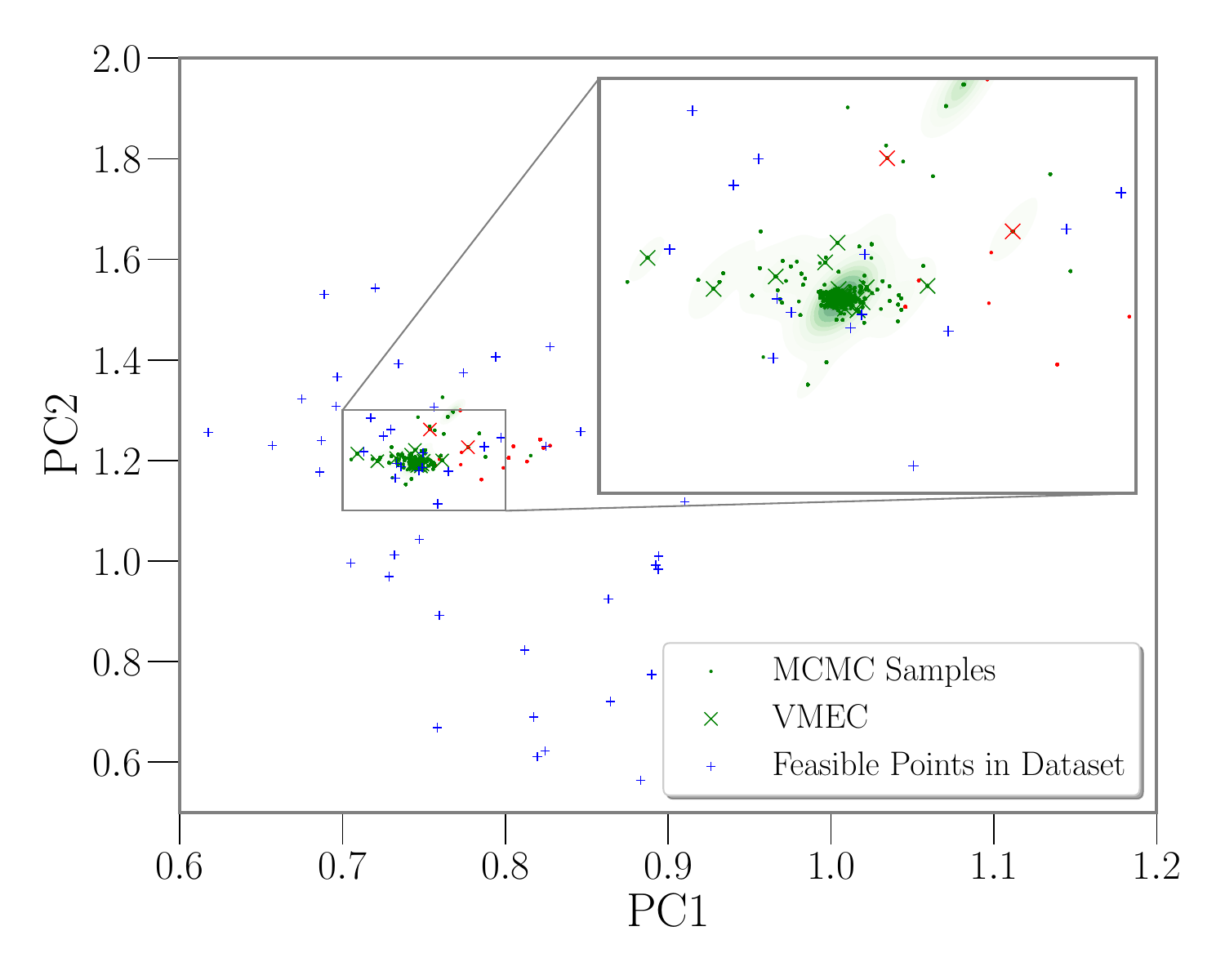}  
    \caption{Simple-to-build problem.}
    \label{fig:posterior-b}
  \end{subfigure}
  \caption{%
    Posterior estimate of the feasible region in the first two \gls{PCA} dimensions for two constraint-relaxed problems.
    Blue crosses represent feasible configurations from the dataset.
    Green dots show \gls{MCMC} samples predicted to be feasible (classifier confidence $\ge \num{0.99}$),
    and red dots indicate predicted infeasible samples.
    Green contours reflect the estimated density of feasible samples.
    Oracle validation of randomly selected \gls{MCMC} points are marked with green (feasible) and red (infeasible) crosses.
    Both the geometric and simple-to-build problems are initially relaxed,
    with $41$ and $52$ feasible points available in the dataset (out of $\sim160k$).
  }
  \label{fig:generative_model_results}
\end{figure}

We reduce the input dimensionality using \gls{PCA}~\cite{abdi2010principal} to obtain a low-dimensional latent space. 
In this space, Random Forest classifiers ~\cite{bishop2006pattern,murphy2023probabilistic} estimate the probability that a configuration is feasible. 
Thresholding this probability (\eg, above 0.8) defines a soft feasible region. 
Within this region, we fit a \gls{GMM} to capture the distribution of feasible points.
Treating the \gls{GMM} as a prior and the classifier output as a quasi-likelihood, we use adaptive \gls{MCMC} \cite{helander2009bootstrap,caflisch1998monte} to sample from the posterior. 
This allows us to generate several new configurations that are likely to satisfy constraints without querying the oracle
(~\cref{fig:generative_model_results}). 
Details are given in~\cref{app:gen_model_appendix} with full algorithmic details in~\cref{alg:feasible_generation}.

When applied to relaxed versions of both the Geometric and Simple-to-build problems, our method successfully identifies regions of design space in which sample points are judged feasible by both the Random Forest classifier and the oracle model (\ie, using \vmecpp) (\cref{fig:generative_model_results}).

\section{Discussion}
\label{sec:discussion}
We released a diverse dataset of approximately $158k$ \gls{QI}-like stellarator plasma boundaries,
associated metrics,
and ideal-\gls{MHD} equilibria in vacuum and at five levels of plasma beta values.
Alongside the dataset,
we introduced a set of stellarator optimization tasks with strong classical baselines,
designed to facilitate rigorous and reproducible evaluation of stellarator optimization strategies.
We further demonstrated a data-driven generative approach that can produce feasible plasma configurations without querying an expensive physics oracle.
Nonetheless, several limitations remain.
First,
the degree of \gls{QI} in the dataset is inherently limited by the finite-budget,
optimization-based sampling process used during generation.
Second,
the dataset is limited to plasma boundaries;
while these are usually the seeds in stellarator design,
a consistent design also requires many additional systems (\eg, electromagnetic coils).

\section*{Acknowledgments}

This work was independently funded by Proxima Fusion,
and supported by the BMBF grant FUSKI (FKZ: 13F1012A).

\bibliographystyle{unsrtnat}
\bibliography{main}

@article{schilling2025numerics,
  title={The Numerics of VMEC++},
  author={Schilling, Jonathan},
  journal={arXiv preprint arXiv:2502.04374},
  year={2025}
}

@article{dudt2020desc,
  title={DESC: a stellarator equilibrium solver},
  author={Dudt, DW and Kolemen, E},
  journal={Physics of Plasmas},
  volume={27},
  number={10},
  year={2020},
  publisher={AIP Publishing}
}

@article{ragonneau2022model,
  title={Model-based derivative-free optimization methods and software},
  author={Ragonneau, Tom M},
  journal={arXiv preprint arXiv:2210.12018},
  year={2022}
}

@article{hestenes1969multiplier,
  title={Multiplier and gradient methods},
  author={Hestenes, Magnus R},
  journal={Journal of optimization theory and applications},
  volume={4},
  number={5},
  pages={303--320},
  year={1969},
  publisher={Springer}
}

@article{powell1969method,
  title={A method for nonlinear constraints in minimization problems},
  author={Powell, Michael JD},
  journal={Optimization},
  pages={283--298},
  year={1969},
  publisher={Academic Press}
}

@article{rockafellar1976augmented,
  title={Augmented Lagrangians and applications of the proximal point algorithm in convex programming},
  author={Rockafellar, R Tyrrell},
  journal={Mathematics of operations research},
  volume={1},
  number={2},
  pages={97--116},
  year={1976},
  publisher={INFORMS}
}

@article{laude2023anisotropic,
  title={Anisotropic proximal point algorithm},
  author={Laude, Emanuel and Patrinos, Panagiotis},
  journal={arXiv preprint arXiv:2312.09834},
  year={2023}
}

@article{Henneberg_2021, 
title={Combined plasma–coil optimization algorithms}, 
volume={87}, 
DOI={10.1017/S0022377821000271}, 
number={2}, 
journal={Journal of Plasma Physics}, 
author={Henneberg, S. A. and Hudson, S. R. and Pfefferlé, D. and Helander, P.}, 
year={2021}, 
pages={905870226}
}

@article{10.1063/1.4765691,
    author = {Hudson, S. R. and Dewar, R. L. and Dennis, G. and Hole, M. J. and McGann, M. and von Nessi, G. and Lazerson, S.},
    title = {Computation of multi-region relaxed magnetohydrodynamic equilibria},
    journal = {Physics of Plasmas},
    volume = {19},
    number = {11},
    pages = {112502},
    year = {2012},
    month = {11},
    issn = {1070-664X},
    doi = {10.1063/1.4765691},
    url = {https://doi.org/10.1063/1.4765691},
}

@article{kappel2024magnetic,
  title={The magnetic gradient scale length explains why certain plasmas require close external magnetic coils},
  author={Kappel, John and Landreman, Matt and Malhotra, Dhairya},
  journal={Plasma Physics and Controlled Fusion},
  volume={66},
  number={2},
  pages={025018},
  year={2024},
  publisher={IOP Publishing}
}

@article{hegna2025infinity,
  title={The infinity two fusion pilot plant baseline plasma physics design},
  author={Hegna, CC and Anderson, DT and Andrew, EC and Ayilaran, A and Bader, A and Bohm, TD and Mata, K Camacho and Canik, JM and Carbajal, L and Cerfon, A and others},
  journal={Journal of Plasma Physics},
  pages={1--44},
  year={2025},
  publisher={Cambridge University Press}
}

@article{lion2025stellaris,
  title={Stellaris: A high-field quasi-isodynamic stellarator for a prototypical fusion power plant},
  author={Lion, J and Angl{\`e}s, J-C and Bonauer, L and Navarro, A Ba{\~n}{\'o}n and Ceron, SA Cadena and Davies, R and Drevlak, M and Foppiani, N and Geiger, J and Goodman, A and others},
  journal={Fusion Engineering and Design},
  volume={214},
  pages={114868},
  year={2025},
  publisher={Elsevier}
}

@article{wiedman2023coil,
  title={Coil optimization for quasi-helically symmetric stellarator configurations},
  author={Wiedman, Alexander Vyacheslav and Buller, Stefan and Landreman, Matt},
  journal={arXiv preprint arXiv:2311.16386},
  year={2023}
}

@article{wechsung2022precise,
  title={Precise stellarator quasi-symmetry can be achieved with electromagnetic coils},
  author={Wechsung, Florian and Landreman, Matt and Giuliani, Andrew and Cerfon, Antoine and Stadler, Georg},
  journal={Proceedings of the National Academy of Sciences},
  volume={119},
  number={13},
  pages={e2202084119},
  year={2022},
  publisher={National Academy of Sciences}
}

@article{liu2018magnetic,
  title={Magnetic configuration and modular coil design for the Chinese first quasi-axisymmetric stellarator},
  author={Liu, Haifeng and Shimizu, Akihiro and Isobe, Mitsutaka and Okamura, Shoichi and Nishimura, Shin and Suzuki, Chihiro and Xu, Yuhong and Zhang, Xin and Liu, Bing and Huang, Jie and others},
  journal={Plasma and Fusion Research},
  volume={13},
  pages={3405067--3405067},
  year={2018},
  publisher={The Japan Society of Plasma Science and Nuclear Fusion Research}
}

@article{jorge2022single,
  title={A single-field-period quasi-isodynamic stellarator},
  author={Jorge, Rogerio and Plunk, GG and Drevlak, M and Landreman, M and Lobsien, J-F and Mata, K Camacho and Helander, P},
  journal={Journal of Plasma Physics},
  volume={88},
  number={5},
  pages={175880504},
  year={2022},
  publisher={Cambridge University Press}
}

@article{goodman2023constructing,
  title={Constructing precisely quasi-isodynamic magnetic fields},
  author={Goodman, Alan G and Mata, K Camacho and Henneberg, Sophia A and Jorge, Rogerio and Landreman, Matt and Plunk, GG and Smith, HM and Mackenbach, RJJ and Beidler, CD and Helander, P},
  journal={Journal of Plasma Physics},
  volume={89},
  number={5},
  pages={905890504},
  year={2023},
  publisher={Cambridge University Press}
}

@article{spears1980scaling,
  title={Scaling of tokamak reactor costs},
  author={Spears, WR and Wesson, JA},
  journal={Nuclear Fusion},
  volume={20},
  number={12},
  pages={1525},
  year={1980},
  publisher={IOP Publishing}
}

@article{freidberg2015designing,
  title={Designing a tokamak fusion reactor—How does plasma physics fit in?},
  author={Freidberg, JP and Mangiarotti, FJ and Minervini, J},
  journal={Physics of Plasmas},
  volume={22},
  number={7},
  year={2015},
  publisher={AIP Publishing}
}

@article{mercier1964equilibrium,
  title={Equilibrium and stability of a toroidal magnetohydrodynamic system in the neighbourhood of a magnetic axis},
  author={Mercier, Claude},
  journal={Nuclear Fusion},
  volume={4},
  number={3},
  pages={213},
  year={1964},
  publisher={IOP Publishing}
}

@article{greene1997brief,
  title={A brief review of magnetic wells},
  author={Greene, John M},
  journal={Comments on Plasma Physics and Controlled Fusion},
  volume={17},
  pages={389--402},
  year={1997},
  publisher={Citeseer}
}

@article{landreman2025does,
  title={How does ion temperature gradient turbulence depend on magnetic geometry? Insights from data and machine learning},
  author={Landreman, Matt and Choi, Jong Youl and Alves, Caio and Balaprakash, Prasanna and Churchill, R Michael and Conlin, Rory and Roberg-Clark, Gareth},
  journal={arXiv preprint arXiv:2502.11657},
  year={2025}
}

@article{roberg2024reduction,
  title={Reduction of electrostatic turbulence in a quasi-helically symmetric stellarator via critical gradient optimization},
  author={Roberg-Clark, GT and Xanthopoulos, P and Plunk, GG},
  journal={Journal of Plasma Physics},
  volume={90},
  number={3},
  pages={175900301},
  year={2024},
  publisher={Cambridge University Press}
}

@article{goodman2024quasi,
  title={Quasi-isodynamic stellarators with low turbulence as fusion reactor candidates},
  author={Goodman, Alan G and Xanthopoulos, Pavlos and Plunk, Gabriel G and Smith, H{\aa}kan and N{\"u}hrenberg, Carolin and Beidler, Craig D and Henneberg, Sophia A and Roberg-Clark, Gareth and Drevlak, Michael and Helander, Per},
  journal={PRX Energy},
  volume={3},
  number={2},
  pages={023010},
  year={2024},
  publisher={APS}
}

@article{beurskens2021ion,
  title={Ion temperature clamping in Wendelstein 7-X electron cyclotron heated plasmas},
  author={Beurskens, Marc NA and Bozhenkov, Sergey A and Ford, O and Xanthopoulos, Pavlos and Zocco, Alessandro and Turkin, Yuri and Alonso, A and Beidler, Craig and Calvo, I and Carralero, Daniel and others},
  journal={Nuclear Fusion},
  volume={61},
  number={11},
  pages={116072},
  year={2021},
  publisher={IOP Publishing}
}

@article{beidler2021demonstration,
  title={Demonstration of reduced neoclassical energy transport in Wendelstein 7-X},
  author={Beidler, CD and Smith, HM and Alonso, A and Andreeva, T and Baldzuhn, J and Beurskens, MNA and Borchardt, Matthias and Bozhenkov, SA and Brunner, Kai Jakob and Damm, Hannes and others},
  journal={Nature},
  volume={596},
  number={7871},
  pages={221--226},
  year={2021},
  publisher={Nature Publishing Group UK London}
}

@article{li2019quality,
  title={Quality evaluation of solution sets in multiobjective optimisation: A survey},
  author={Li, Miqing and Yao, Xin},
  journal={ACM Computing Surveys (CSUR)},
  volume={52},
  number={2},
  pages={1--38},
  year={2019},
  publisher={ACM New York, NY, USA}
}

@inproceedings{zitzler1998multiobjective,
  title={Multiobjective optimization using evolutionary algorithms—a comparative case study},
  author={Zitzler, Eckart and Thiele, Lothar},
  booktitle={International conference on parallel problem solving from nature},
  pages={292--301},
  year={1998},
  organization={Springer}
}

@article{rummel2004accuracy,
  title={Accuracy of the construction of the superconducting coils for Wendelstein 7-X},
  author={Rummel, Thomas and Risse, Konrad and Viebke, Holger and Braeuer, Torsten and Kisslinger, Johann},
  journal={IEEE transactions on applied superconductivity},
  volume={14},
  number={2},
  pages={1394--1398},
  year={2004},
  publisher={IEEE}
}

@article{neilson2010lessons,
  title={Lessons learned in risk management on NCSX},
  author={Neilson, GH and Gruber, CO and Harris, Jeffrey H and Rej, DJ and Simmons, RT and Strykowsky, RL},
  journal={IEEE transactions on plasma science},
  volume={38},
  number={3},
  pages={320--327},
  year={2010},
  publisher={IEEE}
}

@techreport{NCSXCloseout2009,
  title        = {National Compact Stellarator Experiment (NCSX) Project Closeout Report},
  author       = {},
  institution  = {Princeton Plasma Physics Laboratory},
  year         = {2009},
  month        = {August},
  url          = {https://ncsx.pppl.gov/NCSX_Engineering/CloseOut_Documentation/Closeout_Report_20080910.pdf},
  note         = {Prepared for the U.S. Department of Energy Office of Science}
}

@article{helander2009bootstrap,
  title={Bootstrap current and neoclassical transport in quasi-isodynamic stellarators},
  author={Helander, Per and N{\"u}hrenberg, J},
  journal={Plasma Physics and Controlled Fusion},
  volume={51},
  number={5},
  pages={055004},
  year={2009},
  publisher={IOP Publishing}
}

@techreport{hirshman1983steepest,
  title={Steepest descent moment method for three-dimensional magnetohydrodynamic equilibria},
  author={Hirshman, Steven P and Whitson, JC},
  year={1983},
  institution={Oak Ridge National Lab.(ORNL), Oak Ridge, TN (United States)}
}

@article{landreman2021simsopt,
  title={SIMSOPT: a flexible framework for stellarator optimization},
  author={Landreman, Matt and Medasani, Bharat and Wechsung, Florian and Giuliani, Andrew and Jorge, Rogerio and Zhu, Caoxiang},
  journal={Journal of Open Source Software},
  volume={6},
  number={65},
  pages={3525},
  year={2021}
}

@article{dudt2024magnetic,
  title={Magnetic fields with general omnigenity},
  author={Dudt, Daniel W and Goodman, Alan G and Conlin, Rory and Panici, Dario and Kolemen, Egemen},
  journal={Journal of Plasma Physics},
  volume={90},
  number={1},
  pages={905900120},
  year={2024},
  publisher={Cambridge University Press}
}

@article{curvo2025using,
  title={Using deep learning to design high aspect ratio fusion devices},
  author={Curvo, P and Ferreira, DR and Jorge, R},
  journal={Journal of Plasma Physics},
  volume={91},
  number={1},
  pages={E38},
  year={2025},
  publisher={Cambridge University Press}
}

@article{landreman2022mapping,
  title={Mapping the space of quasisymmetric stellarators using optimized near-axis expansion},
  author={Landreman, Matt},
  journal={Journal of Plasma Physics},
  volume={88},
  number={6},
  pages={905880616},
  year={2022},
  publisher={Cambridge University Press}
}

@article{giuliani2024direct,
  title={Direct stellarator coil design using global optimization: application to a comprehensive exploration of quasi-axisymmetric devices},
  author={Giuliani, Andrew},
  journal={Journal of Plasma Physics},
  volume={90},
  number={3},
  pages={905900303},
  year={2024},
  publisher={Cambridge University Press}
}

@article{giuliani2024comprehensive,
  title={A comprehensive exploration of quasisymmetric stellarators and their coil sets},
  author={Giuliani, Andrew and Rodr{\'\i}guez, Eduardo and Spivak, Marina},
  journal={arXiv preprint arXiv:2409.04826},
  year={2024}
}

@article{conlin2024stellarator,
  title={Stellarator optimization with constraints},
  author={Conlin, Rory and Kim, Patrick and Dudt, Daniel W and Panici, Dario and Kolemen, Egemen},
  journal={Journal of Plasma Physics},
  volume={90},
  number={5},
  pages={905900501},
  year={2024},
  publisher={Cambridge University Press}
}

@article{nies2024exploration,
  title={Exploration of the parameter space of quasisymmetric stellarator vacuum fields through adjoint optimisation},
  author={Nies, Richard and Paul, Elizabeth J and Panici, Dario and Hudson, Stuart R and Bhattacharjee, Amitava},
  journal={Journal of Plasma Physics},
  volume={90},
  number={6},
  pages={905900620},
  year={2024},
  publisher={Cambridge University Press}
}

@article{buller2025family,
  title={A family of quasi-axisymmetric stellarators with varied rotational transform},
  author={Buller, Stefan and Landreman, Matt and Kappel, John and Gaur, Rahul},
  journal={Journal of Plasma Physics},
  volume={91},
  number={1},
  pages={E18},
  year={2025},
  publisher={Cambridge University Press}
}

@article{garren1991magnetic,
  title={Magnetic field strength of toroidal plasma equilibria},
  author={Garren, David Alan and Boozer, AH},
  journal={Physics of Fluids B: Plasma Physics},
  volume={3},
  number={10},
  pages={2805--2821},
  year={1991},
  publisher={American Institute of Physics}
}

@article{garren1991existence,
  title={Existence of quasihelically symmetric stellarators},
  author={Garren, DA and Boozer, Allen H},
  journal={Physics of Fluids B},
  volume={3},
  number={10},
  pages={2822--2834},
  year={1991}
}

@article{landreman2019optimized,
  title={Optimized quasisymmetric stellarators are consistent with the Garren--Boozer construction},
  author={Landreman, Matt},
  journal={Plasma Physics and Controlled Fusion},
  volume={61},
  number={7},
  pages={075001},
  year={2019},
  publisher={IOP Publishing}
}

@article{landreman2019constructing,
  title={Constructing stellarators with quasisymmetry to high order},
  author={Landreman, Matt and Sengupta, Wrick},
  journal={Journal of Plasma Physics},
  volume={85},
  number={6},
  pages={815850601},
  year={2019},
  publisher={Cambridge University Press}
}

@article{rodriguez2023constructing,
  title={Constructing the space of quasisymmetric stellarators through near-axis expansion},
  author={Rodr{\'\i}guez, Eduardo and Sengupta, Wrick and Bhattacharjee, Amitava},
  journal={Plasma Physics and Controlled Fusion},
  volume={65},
  number={9},
  pages={095004},
  year={2023},
  publisher={IOP Publishing}
}

@article{henneberg2021representing,
  title={Representing the boundary of stellarator plasmas},
  author={Henneberg, Sophia A and Helander, Per and Drevlak, Michael},
  journal={Journal of Plasma Physics},
  volume={87},
  number={5},
  pages={905870503},
  year={2021},
  publisher={Cambridge University Press}
}

@misc{hardt2025emerging,
  author = {Moritz Hardt},
  title = {The Emerging Science of Machine Learning Benchmarks},
  year = {2025},
  howpublished = {Online at \url{https://mlbenchmarks.org}},
  note = {Manuscript}
}

@techreport{boozer1981plasma,
  title={Plasma equilibrium with rational magnetic surfaces},
  author={Boozer, Allen H},
  year={1981},
  institution={Princeton Plasma Physics Lab.(PPPL), Princeton, NJ (United States)}
}

@book{wesson2011tokamaks,
  title={Tokamaks},
  author={Wesson, John and Campbell, David J},
  volume={149},
  year={2011},
  publisher={Oxford university press}
}

@article{hender2007mhd,
  title={MHD stability, operational limits and disruptions},
  author={Hender, TC and Wesley, JC and Bialek, J and Bondeson, A and Boozer, AH and Buttery, RJ and Garofalo, A and Goodman, TP and Granetz, RS and Gribov, Y and others},
  journal={Nuclear fusion},
  volume={47},
  number={6},
  pages={S128},
  year={2007},
  publisher={IOP Publishing}
}

@book{freidberg2014ideal,
  title={ideal MHD},
  author={Freidberg, Jeffrey P},
  year={2014},
  publisher={Cambridge University Press}
}

@book{bishop2006pattern,
  title={Pattern recognition and machine learning},
  author={Bishop, Christopher M and Nasrabadi, Nasser M},
  volume={4},
  number={4},
  year={2006},
  publisher={Springer}
}

@book{murphy2023probabilistic,
  title={Probabilistic machine learning: Advanced topics},
  author={Murphy, Kevin P},
  year={2023},
  publisher={MIT press}
}

@misc{blondel_elements_2024,
	title = {The {Elements} of {Differentiable} {Programming}},
	url = {http://arxiv.org/abs/2403.14606},
	doi = {10.48550/arXiv.2403.14606},
	urldate = {2024-03-25},
	publisher = {arXiv},
	author = {Blondel, Mathieu and Roulet, Vincent},
	month = mar,
	year = {2024},
	note = {arXiv:2403.14606 [cs]},
}

@article{haario2001adaptive,
  title={An adaptive Metropolis algorithm},
  author={Haario, Heikki and Saksman, Eero and Tamminen, Johanna},
  journal={Bernoulli},
  pages={223--242},
  year={2001},
  publisher={JSTOR}
}

@article{caflisch1998monte,
  title={{M}onte {C}arlo and quasi-{M}onte {C}arlo methods},
  author={Caflisch, Russel E},
  journal={Acta numerica},
  volume={7},
  pages={1--49},
  year={1998},
  publisher={Cambridge University Press}
}

@article{hastings1970monte,
  title={{M}onte {C}arlo sampling methods using Markov chains and their applications},
  author={Hastings, W Keith},
  year={1970},
  publisher={Oxford University Press}
}

@techreport{baker2019workshop,
  title={Workshop report on basic research needs for scientific machine learning: Core technologies for artificial intelligence},
  author={Baker, Nathan and Alexander, Frank and Bremer, Timo and Hagberg, Aric and Kevrekidis, Yannis and Najm, Habib and Parashar, Manish and Patra, Abani and Sethian, James and Wild, Stefan and others},
  year={2019},
  institution={USDOE Office of Science (SC), Washington, DC (United States)}
}

@article{cuomo2022scientific,
  title={Scientific machine learning through physics--informed neural networks: Where we are and what’s next},
  author={Cuomo, Salvatore and Di Cola, Vincenzo Schiano and Giampaolo, Fabio and Rozza, Gianluigi and Raissi, Maziar and Piccialli, Francesco},
  journal={Journal of Scientific Computing},
  volume={92},
  number={3},
  pages={88},
  year={2022},
  publisher={Springer}
}

@article{abdi2010principal,
  title={Principal component analysis},
  author={Abdi, Herv{\'e} and Williams, Lynne J},
  journal={Wiley interdisciplinary reviews: computational statistics},
  volume={2},
  number={4},
  pages={433--459},
  year={2010},
  publisher={Wiley Online Library}
}

@article{shirobokov2020black,
  title={Black-box optimization with local generative surrogates},
  author={Shirobokov, Sergey and Belavin, Vladislav and Kagan, Michael and Ustyuzhanin, Andrei and Baydin, Atilim Gunes},
  journal={Advances in Neural Information Processing Systems},
  volume={33},
  pages={14650--14662},
  year={2020}
}

@book{martins2021engineering,
  title={Engineering design optimization},
  author={Martins, Joaquim RRA and Ning, Andrew},
  year={2021},
  publisher={Cambridge University Press}
}

@article{scikit-learn,
  title={Scikit-learn: Machine Learning in {P}ython},
  author={Pedregosa, F. and Varoquaux, G. and Gramfort, A. and Michel, V.
          and Thirion, B. and Grisel, O. and Blondel, M. and Prettenhofer, P.
          and Weiss, R. and Dubourg, V. and Vanderplas, J. and Passos, A. and
          Cournapeau, D. and Brucher, M. and Perrot, M. and Duchesnay, E.},
  journal={Journal of Machine Learning Research},
  volume={12},
  pages={2825--2830},
  year={2011}
}

@article{helander2014theory,
  title={Theory of plasma confinement in non-axisymmetric magnetic fields},
  author={Helander, Per},
  journal={Reports on Progress in Physics},
  volume={77},
  number={8},
  pages={087001},
  year={2014},
  publisher={IOP Publishing}
}

@article{cary1997omnigenity,
  title={Omnigenity and quasihelicity in helical plasma confinement systems},
  author={Cary, John R and Shasharina, Svetlana G},
  journal={Physics of Plasmas},
  volume={4},
  number={9},
  pages={3323--3333},
  year={1997},
  publisher={American Institute of Physics}
}

@article{jorge2020near,
  title={Near-axis expansion of stellarator equilibrium at arbitrary order in the distance to the axis},
  author={Jorge, R and Sengupta, W and Landreman, M},
  journal={Journal of Plasma Physics},
  volume={86},
  number={1},
  pages={905860106},
  year={2020},
  publisher={Cambridge University Press}
}

@article{byrd1999interior,
  title={An interior point algorithm for large-scale nonlinear programming},
  author={Byrd, Richard H and Hribar, Mary E and Nocedal, Jorge},
  journal={SIAM Journal on Optimization},
  volume={9},
  number={4},
  pages={877--900},
  year={1999},
  publisher={SIAM}
}

@misc{nevergrad,
    author = {J. Rapin and O. Teytaud},
    title = {{Nevergrad - A gradient-free optimization platform}},
    year = {2018},
    publisher = {GitHub},
    journal = {GitHub repository},
    howpublished = {\url{https://GitHub.com/FacebookResearch/Nevergrad}},
}

@inproceedings{dufosse2021augmented,
  title={Augmented Lagrangian, penalty techniques and surrogate modeling for constrained optimization with CMA-ES},
  author={Dufoss{\'e}, Paul and Hansen, Nikolaus},
  booktitle={Proceedings of the Genetic and Evolutionary Computation Conference},
  pages={519--527},
  year={2021}
}

\newpage
\appendix

\section{Technical Appendices}

\subsection{Data generation: sampling omnigenous poloidal fields}
\label{app:sampling}

We leverage the parameterization of an \textit{omnigenous poloidal} field from \citet{dudt2024magnetic} in which the 1D magnetic well on each flux surface is represented by a spline on the interval $[-\pi/2, \pi/2]$.
The well is symmetric about its minimum, so it can simply be parameterized between $B_{min}$ and $B_{max}$. 
The full omnigenous field is then built by ``morphing" this one-dimensional well across magnetic field lines via a computational coordinate $h$,
which is expanded in a Chebyshev basis (radial index $l$) and Fourier bases (poloidal index $m$, toroidal index $n$) with coefficients $x_{lmn}$ \cite{dudt2024magnetic}. 
In practice, we generate new omnigenous-poloidal fields by sampling both the spline knots and $x_{lmn}$ coefficients.
To enforce stellarator symmetry (invariance under simultaneous flips of the poloidal and toroidal Boozer angles), we set the coefficients of the odd terms of the Fourier basis along the toroidal direction to zero, namely $x_{lmn} = 0 \ \forall n >= 0$.

To produce a variety of monotonically increasing 1D well shapes, we draw knot positions from $\mathrm{Beta}(\alpha, \beta)$ cumulative distribution functions and then rescaled them to lie between $B_{min}$ and $B_{max}$. 
Finally, we fix the mean magnetic field at 1 T and sample the mirror ratio $\Delta$ to determine the pair $(B_{min}, B_{max})$. 

The ranges from which we sample these parameters can be found in Table \ref{tab:sampling_ramges}.

\begin{table}[htbp]
  \centering
  \label{tab:sampling_ranges}
  \footnotesize
  \begin{tabular}{@{}lcc@{}}
    \toprule
    Parameter & Min & Max \\
    \midrule
    $N_{\mathit{fp}}$ & 1 & 5 \\
    $\tilde\iota$ & 0.1 & 0.3 \\
    $A$            & 4.0 & 12.0 \\
    $\epsilon_{\mathrm{max}}$          & 4.0 & 7.0  \\
    $\alpha_{\mathit{Beta}}$        & 2.0 & 6.0  \\
    $\beta_{\mathit{Beta}}$      & 2.0 & 6.0  \\
    $\Delta_{\mathit{edge}}$ & 0.1 & 0.4  \\
    \bottomrule
  \end{tabular}
  \caption{Ranges of sampling parameters with both minimum and maximum values. }
  \label{tab:sampling_ramges}
\end{table}

\subsection{Data generation: stage one optimizations using \texttt{DESC} \cite{dudt2020desc}}
\label{app:desc_optimization}

Given a set of target quantities:
\[
T = \bigl(\iota^*,\,A^*,\,E^*,\,\mathcal{O}^*\bigr)
\]
where
\begin{itemize}
  \item $\iota^*$ is the desired edge rotational transform,
  \item $A^*$ is the target aspect ratio,
  \item $E^*$ is the maximum elongation,
  \item $\mathcal{O}^*$ is the target omnigenous field,
\end{itemize}

we ran numerical optimizations to find a toroidal boundary surface~$\Sigma$ (parameterized in a Fourier‐$RZ$ basis) that simultaneously matches these goals. Note that the mirror ratio $\Delta$ is defined within $\mathcal{O}^*$.

\subsubsection{Initial Guess Generation}
An initial boundary $\Sigma_0$ is generated either by
\begin{enumerate}
  \item \textbf{Heuristic QP model (Section III from \citet{goodman2023constructing})}: prescribing average major radius~$R_0$, aspect ratio $A^*$, elongation $E^*$, mirror ratio, torsion, and field periods; or
  \item \textbf{Near‐Axis Expansion (NAE) using \texttt{pyQSC} \footnote{\url{https://github.com/rogeriojorge/pyQIC}}}: specifying~$A^*$, $E^*$, $\iota^*$, mirror ratio, field periods, and mode cutoffs.
\end{enumerate}
This yields a smooth $\Sigma_0$ expressed in the \texttt{FourierRZToroidalSurface} format of \texttt{DESC}.

\subsubsection{Equilibrium Solve}
Starting from $\Sigma_0$, we form the \texttt{DESC} equilibrium object and solve the force balance
\[
\mathcal{E}(\Sigma) = \text{Equilibrium}\bigl(\Psi,\,\Sigma,\,M,\,N\bigr)
\]
and solve the magnetostatic force‐balance equations using
\[
\mathcal{E} \;\xrightarrow{\ \text{solve}(\text{force})\ }\;\mathcal{E}^\mathrm{sol}\,.
\]

\subsubsection{Objective Function}
On the solved equilibrium~$\mathcal{E}^\mathrm{sol}$, we define individual objective terms:
\begin{align}
  J_{A}(\Sigma) &= \frac{R_0(\Sigma)}{a(\Sigma)}, 
    & f_{A} &= w_{A}\,\bigl(J_{A} - A^*\bigr)^2, \\
  J_{E}(\Sigma) &= \max_{\varphi}\frac{b(\varphi;\Sigma)}{a(\varphi;\Sigma)}, 
    & f_{E} &= w_{E}\,\bigl(J_{E} - E^*\bigr)^2, \\
  J_{\iota}(\Sigma) &= \iota\,[\mathcal{E}^\mathrm{sol}], 
    & f_{\iota} &= w_{\iota}\,\bigl(J_{\iota} - \iota^*\bigr)^2, \\
  \mathbf{J}_{\mathcal{O}}(\Sigma) &= \mathcal{O}\bigl[\mathcal{E}^\mathrm{sol},\,\mathcal{O}^*\bigr], 
    & f_{\mathcal{O}} &= w_{\mathcal{O}}\,\bigl\lVert\mathbf{J}_{\mathcal{O}}\bigr\rVert_2^2,
\end{align}
where $a,b$ are the minor/major half‐axes of the cross‐section, $\varphi$ is the toroidal angle, and the omnigenous residual $\mathbf{J}_{\mathcal{O}}$ is computed by the \texttt{DESC} \texttt{Omnigenity} objective using the target field $\mathcal{O}^*$.

The omnigenity contribution $\bigl\lVert\mathbf{J}_{\mathcal{O}}\bigr\rVert_2^2$ is given by
\[
\bigl\lVert\mathbf{J}_{\mathcal{O}}\bigr\rVert_2^2
=\sum_{i=1}^{N_\eta}\sum_{j=1}^{N_\alpha}
w(\eta_i)\,\bigl[
B_{\rm eq}(\rho_0,\eta_i,\alpha_j)
-\;B^*(\rho_0,\eta_i,\alpha_j)
\bigr]^2,
\]
with the poloidal weight
\[
w(\eta)
=\frac{\eta_{\rm weight}+1}{2}
\;+\;
\frac{\eta_{\rm weight}-1}{2}\,\cos\eta
\quad
\bigl(\text{so }w\equiv1\text{ if }\eta_{\rm weight}=1\bigr),
\]
where \(B_{\rm eq}\) is the field strength of the equilibrium in \((\rho,\eta,\alpha)\) coordinates and \(B^*\) is the perfectly‐omnigenous target field generated by \texttt{OmnigenousField} as in \cite{dudt2024magnetic}. The residuals \(r_{ij}= \sqrt{w(\eta_i)}\bigl(B_{\rm eq}-B^*\bigr)_{ij}\) are evaluated on the same \((\eta,\alpha)\)-grid used by the target field.

On a solved equilibrium~\(\mathcal{E}^\mathrm{sol}\) at a fixed flux surface \(\rho=\rho_0\), we assemble a least‐squares objective
\[
\mathcal{L}(\Sigma) \;=\; f_{A} + f_{E} + f_{\iota} + f_{\mathcal{O}}\,.
\]

Internally, \texttt{DESC} invokes \texttt{JAX} to compute residuals, leveraging automatic differentiation to compute gradients.

The objective is then wrapped in an augmented‐Lagrangian least‐squares optimizer (\texttt{lsq-auglag}) \cite{conlin2024stellarator} to minimize \(\|r\|_2^2\) alongside the other terms.

\subsubsection{Constraints}
To enforce vacuum equilibrium and fix global invariants, the following constraints are imposed:
\[
\begin{aligned}
  &R_{0,0}(\Sigma)\;=\; 1, 
  &\quad&(\texttt{FixBoundaryR}) \\
  &j_\parallel(\Sigma)\;=\;0,
  &\quad&(\texttt{CurrentDensity}) \\
  &p(\Sigma)\;=\;0,
  &\quad&(\texttt{FixPressure}) \\
  &J_\mathrm{tor}(\Sigma)\;=\;0,
  &\quad&(\texttt{FixCurrent}) \\
  &\Psi(\Sigma)\;=\;\text{const.},
  &\quad&(\texttt{FixPsi})
\end{aligned}
\]
where each is implemented via the corresponding \texttt{DESC} linear‐objective wrapper.

\subsubsection{Nonlinear Optimization}
We employ \texttt{DESC}’s \texttt{lsq‐auglag} optimizer \cite{conlin2024stellarator} to solve
\[
\min_{\Sigma}\;\mathcal{L}(\Sigma)
\quad\text{s.t. all linear constraints,}
\]
using automatic differentiation and a trust‐region least‐squares augmented‐Lagrangian scheme.  Iterations continue until convergence (up to 200 iterations by default), yielding the optimized boundary~$\Sigma^*$.

Our exact implementation is available at \url{https://github.com/proximafusion/constellaration}.

\subsection{Stage one optimizations using \vmecpp\cite{schilling2025numerics} in the loop}

We carried out optimizations using the \texttt{NGOpt} algorithm from the \texttt{Nevergrad}~\footnote{https://github.com/facebookresearch/nevergrad} library.
To improve convergence,
we preconditioned the problem using a diagonal scaling matrix as detailed in Section A.2.1.
We parameterized the boundary with up to four poloidal and toroidal Fourier modes and ran the optimization on a single machine equipped with 32 vCPUs and 32GB of RAM.
Each run is allocated a time budget of approximately \SI{1}{\hour}.

The optimization minimizes the following objective function:

\begin{equation}
\begin{aligned}
f(\Theta) =\ 
& \int_0^{2 \pi} \int_0^{\pi / N_{\text{fp}}} \left( B(\theta, \phi) - B^{\ast}(\theta, \phi) \right)^2 \, d\theta \, d\phi \\
& + \int_0^{2 \pi} \left( \max_{\phi} B(\theta, \phi) - B(\theta, \phi = 0) \right)^2 \, d\theta \\
& + \left( \frac{A - A^{\ast}}{A^{\ast}} \right)^2 \\
& + \left( \frac{\iota_{\text{edge}} - \iota_{\text{edge}}^{\ast}}{\iota_{\text{edge}}^{\ast}} \right)^2 \\
& + \left( \max\left(0, \frac{\epsilon_{\text{max}} - \epsilon_{\text{max}}^{\ast}}{\epsilon_{\text{max}}^{\ast}} \right) \right)^2 \ .
\end{aligned}
\end{equation}

where $B$ denotes the magnetic field strength from the ideal-\gls{MHD} equilibrium in Boozer coordinates,
and $B^{\ast}$ represents the target omnigenous magnetic field strength.
The quantities $A$, $\iota_{\text{edge}}$, and $\epsilon_{\text{max}}$ correspond to the aspect ratio,
edge rotational transform,
and maximum elongation,
respectively,
with asterisks denoting their target values.
The additional target on the maxima of the magnetic field strength guides the optimizer towards more \gls{QI} fields.

In the optimization loop,
we used \vmecpp within the forward model.
To speed up the generation of the optimized boundary,
we run \vmecpp at a lower resolution than the one used to score plasma boundaries in optimization benchmarks
(e.g. reduced number of flux surfaces, higher required force tolerance to converge).

Due to the constrained time budget,
the optimization may not fully minimize the objective function but added the desired diversity to the dataset.

Our exact implementation is available at \url{https://github.com/proximafusion/constellaration}.

\subsection{In-domain predictability of the dataset.}\label{app:predictability_dataset}

We filtered the vacuum data for three field period configurations that were also the result of either the \texttt{DESC} or \vmec optimizations. Then we filtered outliers ($0.05\%$ tails) for each of the metrics, resulting in $\sim 23k$ data points. Finally, we split the data into training and test ($20\%$) sets.

To facilitate training, we also Z-scored the output metrics, keeping track of the training set statistics for later inference.

Using Bayesian optimization to sweep over hyperparameters like network depth, width, type of activation, and learning rate; we converged to an ensemble model of ten multi-layer perceptrons (MLPs) with three layers, 256 hidden units, and \texttt{tanh} activations. The MLPs mapped Fourier‐boundary coefficients to target key metrics by minimizing mean squared error.

We obtained fairly good in-domain generalization results in terms of root mean squared error (RMSE), Pearson's correlation coefficient ($R^2$), normalized root mean squared error (NRMSE), and signal-to-noise ratio (SNR) (Table \ref{tab:surrogate_results}).

\begin{table}[ht]
\centering
\caption{Test-set performances of an MLP ensemble model trained to predict target metrics from boundary coefficients}
\label{tab:mlp_metrics}
\begin{tabular}{lcccc}
\toprule
\textbf{Metric} & \textbf{RMSE} & \textbf{$R^2$} & \textbf{NRMSE} & \textbf{SNR} \\
\midrule
aspect\_ratio & 0.090 & 0.997 & 0.050 & 808.54 \\
aspect\_ratio\_over\_edge\_rotational\_transform & 0.581 & 0.993 & 0.080 & 507.73 \\
max\_elongation & 0.161 & 0.994 & 0.072 & 731.78 \\
axis\_rotational\_transform\_over\_n\_field\_periods & 0.006 & 0.994 & 0.071 & 556.74 \\
edge\_rotational\_transform\_over\_n\_field\_periods & 0.006 & 0.997 & 0.052 & 985.74 \\
axis\_magnetic\_mirror\_ratio & 0.010 & 0.974 & 0.159 & 79.72 \\
edge\_magnetic\_mirror\_ratio & 0.013 & 0.989 & 0.102 & 244.02 \\
average\_triangularity & 0.018 & 0.995 & 0.065 & 806.58 \\
vacuum\_well & 0.006 & 0.998 & 0.062 & 1934.06 \\
minimum\_normalized\_magnetic\_gradient\_scale\_length & 0.330 & 0.990 & 0.101 & 342.57 \\
flux\_compression\_in\_regions\_of\_bad\_curvature & 0.034 & 0.990 & 0.059 & 432.36 \\
log\_10\_qi & 0.051 & 0.982 & 0.134 & 132.22 \\
\bottomrule
\end{tabular}
\label{tab:surrogate_results}
\end{table}

We highlight that such surrogate models are prone to extrapolation errors, particularly when queried far from the data distribution they were trained on \cite{shirobokov2020black}. Uncertainty calibration, active learning, and physics informed strategies (among others) could be considered moving forward for effective surrogate-based optimizations \cite{cuomo2022scientific, baker2019workshop}. 

\subsection{Optimization baselines}\label{app:optimization_baselines}
\subsubsection{Implementation details and hyperparameters}
In this section we provide implementation details for the optimization baseline. For the SciPy-based optimizers,
we use default parameters,
and set the maximum number of iterations to a large value.

We implement a variant of the proximal \gls{ALM}~\cite{rockafellar1976augmented} where the quadratic proximal term is replaced by a trust-region constraint. This can be seen as an instance of the anisotropic proximal \gls{ALM}~ \cite{laude2023anisotropic}.
The modification is essential for improving convergence when using evolutionary algorithms (such as NGOpt),
as it restricts the sampling of new candidate solutions to a region around the current iterate \cite{dufosse2021augmented}.

As the degrees of freedom $\Theta$ operate on different scale,
we precondition the problem with a diagonal matrix $\mathrm{diag}(\Lambda)$ where the entries $\Lambda$ decay exponentially.
We define the rescaled variables as
$\widetilde \Theta := \mathrm{diag}(\Lambda)^{-1} \Theta$
and $\tilde f(\widetilde \Theta) := f(\mathrm{diag} (\Lambda)  \widetilde \Theta)$ and $
\tilde c_i(\widetilde \Theta) := c_i(\mathrm{diag} (\Lambda)  \widetilde \Theta)$.
In addition,
we apply a base-10 logarithmic transformation to the QI constraint.

In each iteration,
the algorithm alternates between primal and dual updates.
For each constraint $\tilde c_i$,
it tracks a penalty parameter $\rho_i^k$ and a Lagrange multiplier $y_i^k$. The complete algorithm is given in~\cref{alg:alm}.

\begin{algorithm}[H]
\caption{non-Euclidean proximal augmented Lagrangian method}
\label{alg:alm}
\begin{algorithmic}[1]
\REQUIRE $\Theta^0 \in \mathbb{R}^D$, $\rho^0 \in \mathbb{R}_{++}^m$, $y^0\in \mathbb{R}_+^m$, $\delta_0 >0$, $0<\tau,\gamma<1$, $\sigma > 1$ and $\delta_{\min}, \rho_{\max} > 0$

\FOR{$k \in \{0, 1, \ldots, N\}$}
\STATE Primal update
\begin{equation} \label{eq:primal_update}
\widetilde\Theta^{k+1}=\argmin_{\widetilde\Theta \in B(\widetilde\Theta^k, \delta_k)} ~\tilde f(\widetilde\Theta) + \tfrac{1}{2} \sum_{i=1}^m \tfrac{1}{\rho_i^k} \Big(\max\{0,  y_i^k + \rho_i^k \tilde c_i(\widetilde\Theta^k)\}^2 - (y_i^k)^2\Big)
\end{equation}
\STATE dual update
$$
y_i^{k+1}=\max\{0, y_i^k + \rho_i^k \tilde c_i(\widetilde\Theta^{k+1})\} 
$$
\STATE update penalty parameters
$$
\rho_i^{k+1} = 
    \begin{cases}
    \rho_i^k & \text{if $\tilde c_i(\widetilde\Theta^{k+1}) \leq \tau \tilde c_i(\widetilde\Theta^{k})$} \\
    \min\{\rho_{\max}, \sigma \rho_i^k \}& \text{otherwise.}
    \end{cases}
$$
\STATE decrease trust-region
$$
 \delta_{k+1} = \max\{ \delta_{\min}, \gamma \delta_k\}
$$
\ENDFOR
\end{algorithmic}
\end{algorithm}

For the geometric problem we choose $\rho_i^0=10$, $\rho_{\max}=1e9$, $\delta_0=0.5,\gamma=0.9, \delta_{\min}=0.05, 
\tau=0.8, \sigma=5$. The subproblem~\eqref{eq:primal_update} is solved with NGOpt with a budget of $\min\{20.000, 1500 + k
\cdot 260\}$ forward-model calls.

For the simple-to-build problem we choose $\rho_i^0=10$, $\rho_{\max}=1e9$, $\delta_0=0.5,\gamma=0.95, \delta_{\min}=0.05, 
\tau=0.8, \sigma=5$. The subproblem~\eqref{eq:primal_update} is solved with NGOpt with a budget of $\min\{20.000, 1500 + k
\cdot 260\}$ forward-model calls.

For the MHD-stable problems we choose $\rho_i^0=10$, $\rho_{\max}=1e8$, $\delta_0=0.33,\gamma=0.95, \delta_{\min}=0.05, 
\tau=0.8, \sigma=5$. The subproblem~\eqref{eq:primal_update} is solved with NGOpt with a budget of $\min\{20.000, 1500 + k
\cdot 300\}$ forward-model calls.

For all problems,
$\Theta^0$ is a rotating ellipse configuration.
We optimize up to four poloidal and toroidal Fourier modes,
which results in $D=80$ degrees of freedom.
During the optimization,
we run \vmecpp at low fidelity.

\subsubsection{Additional experimental results}

\begin{figure}[htbp]
  \centering
  \begin{subfigure}[b]{0.485\textwidth}
    \centering
    \includegraphics[width=\textwidth]{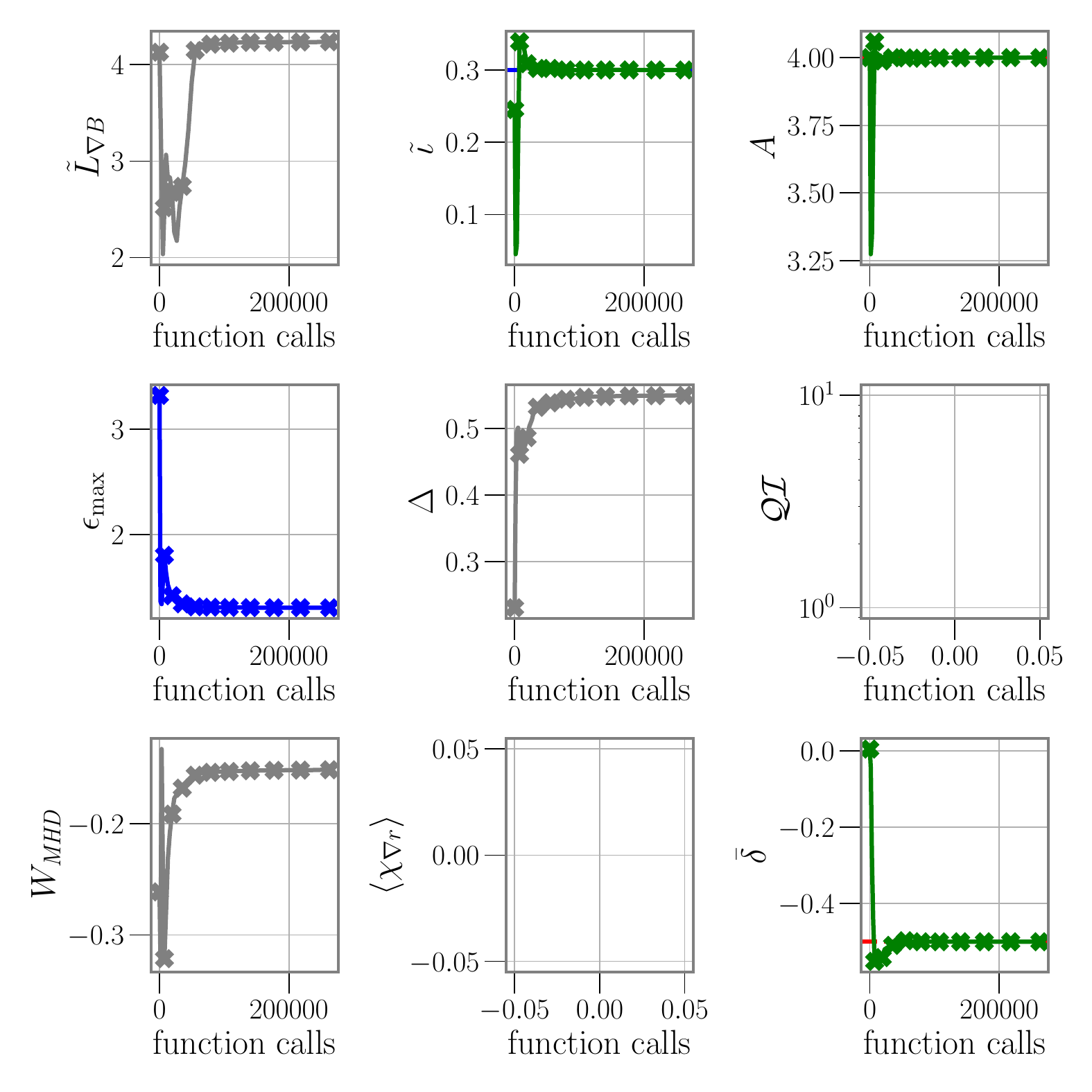}
    \caption{geometric problem.}
    \label{fig:image7}
  \end{subfigure}
  \hfill
  \begin{subfigure}[b]{0.485\textwidth}
    \centering
    \includegraphics[width=\textwidth]{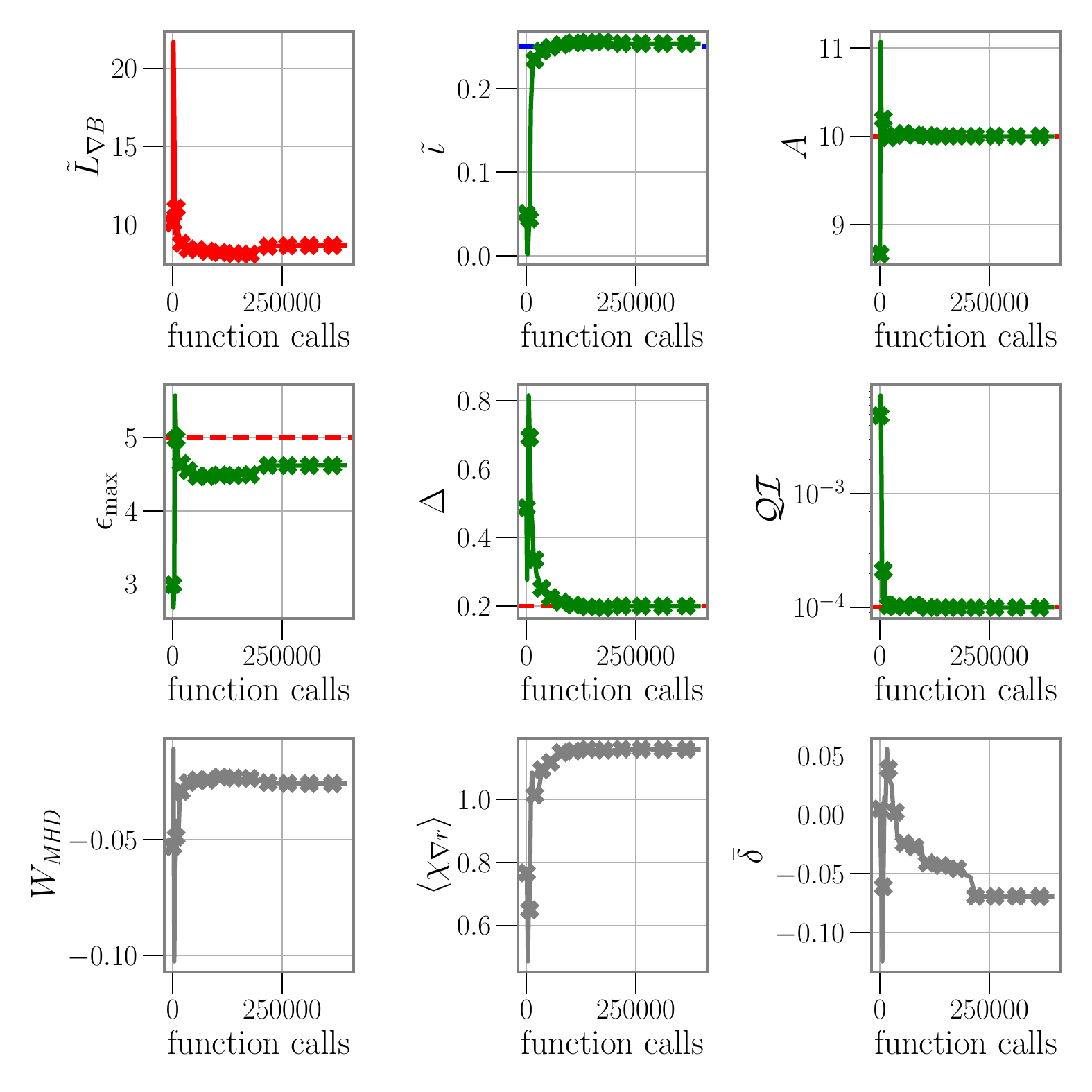}
    \caption{Simple-to-build problem.}
    \label{fig:image8}
  \end{subfigure}
\caption{Single-objective problem optimization traces.}
\label{fig:single_objective}
\end{figure}

\begin{figure}[htbp]
  \centering
  \includegraphics[width=0.485\textwidth]{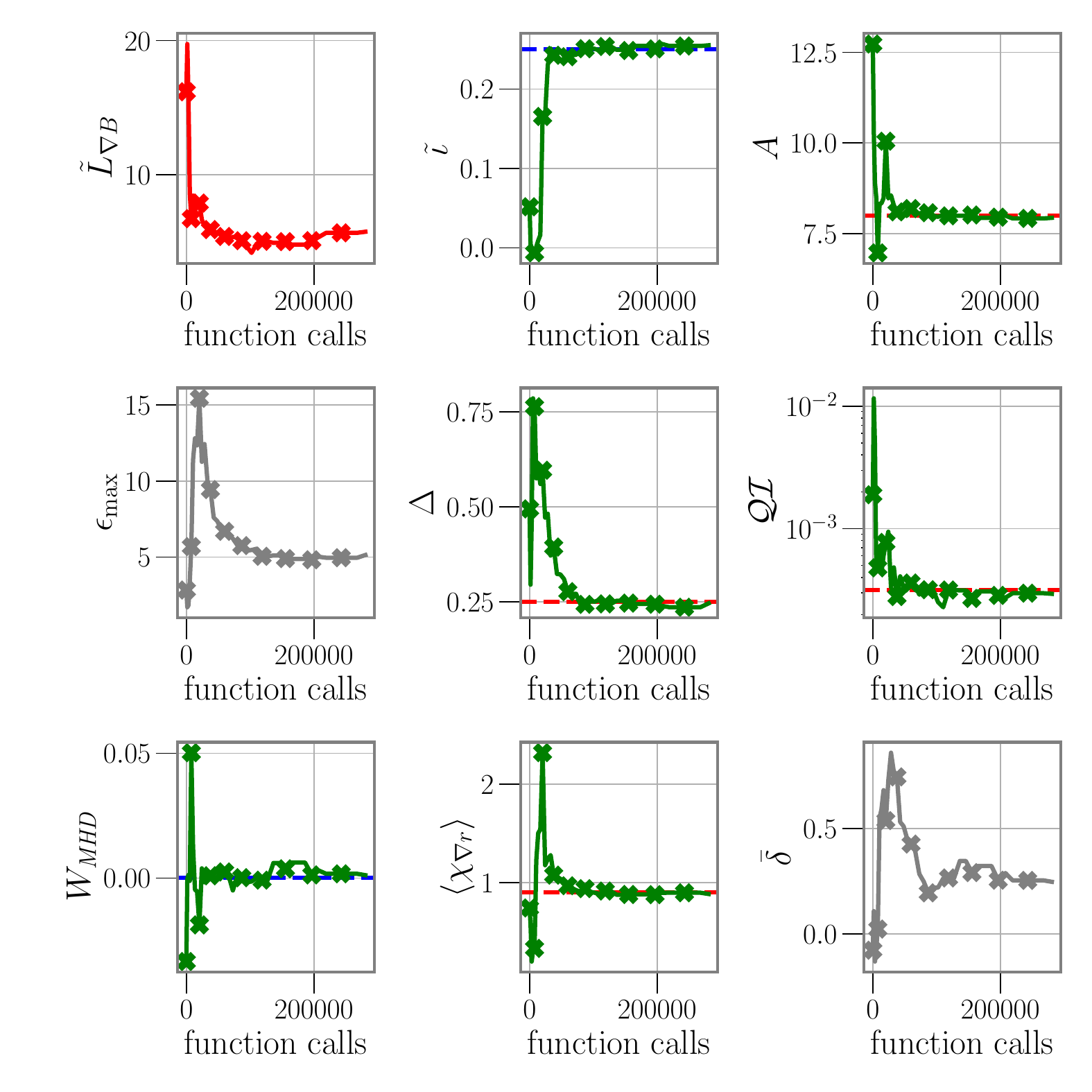}
  \caption{Multi-objective problem with $A\leq 8$.}
\label{fig:multi_objective}
\end{figure}

We provide convergence plots for the three problems obtained with \texttt{ALM-NGOpt}.
Green curves represent metrics that are constrained.
Red colored metrics are maximized and blue colored metrics are minimized.
Gray curves correspond to metrics that are not part of the optimization problem. The blue dashed lines indicate lower bounds and the red dashed lines indicate upper bounds.
\Cref{fig:single_objective} provides plots corresponding to the single-objective problem,
while \cref{fig:multi_objective} provides a plot for one instance ($A\leq 8$) of the sequence of single-objective problems corresponding to the multi-objective problem.

\begin{figure}[htbp]
   \centering
   \begin{subfigure}[b]{0.49\textwidth}
     \centering
     \includegraphics[width=0.45\textwidth]{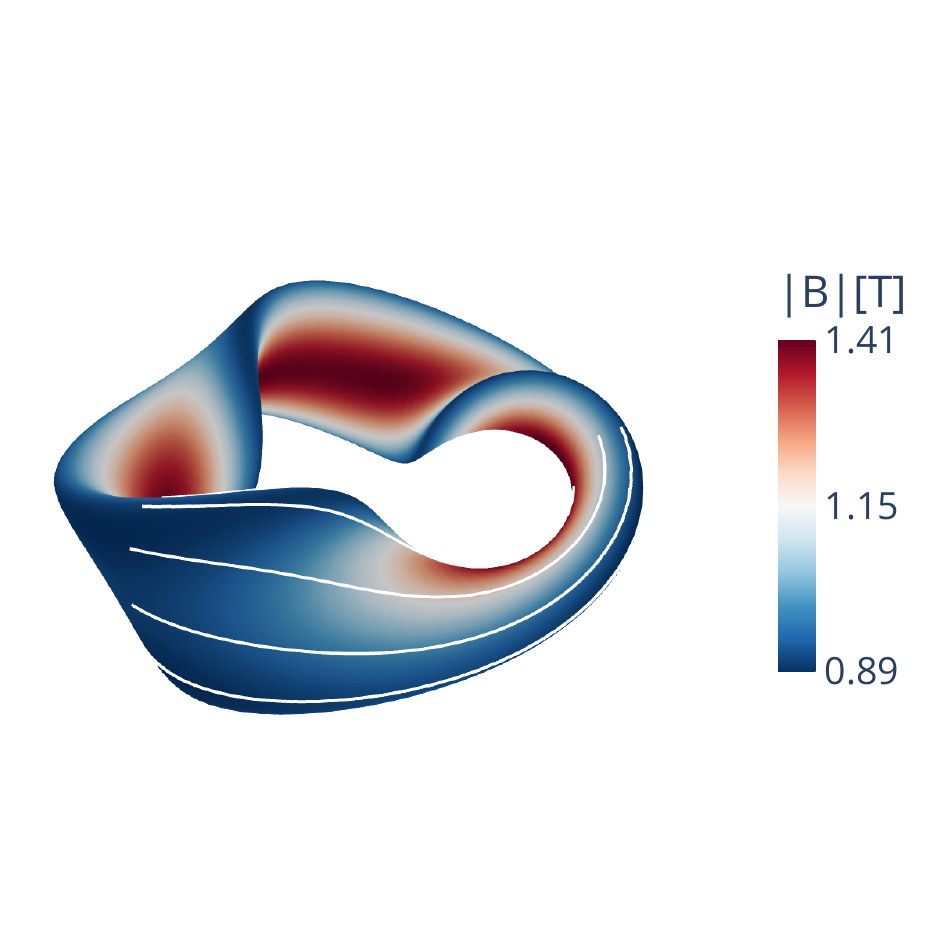}
     \includegraphics[width=0.45\textwidth]{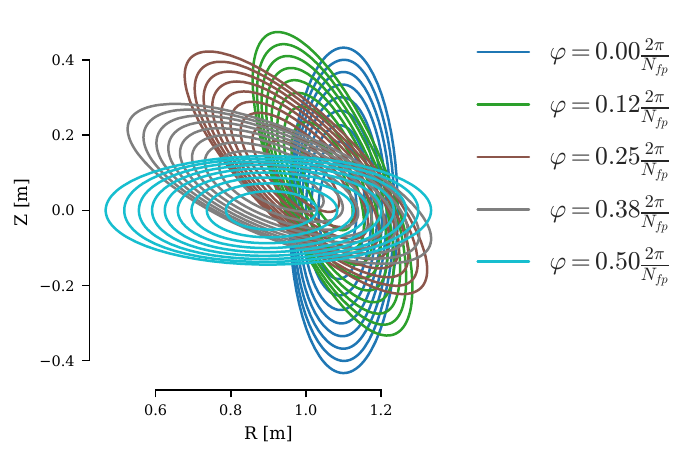}
     \caption{initial geometric}
     \label{fig:image1}
   \end{subfigure}
   \hfill
   \begin{subfigure}[b]{0.49\textwidth}
     \centering
     \includegraphics[width=0.45\textwidth]{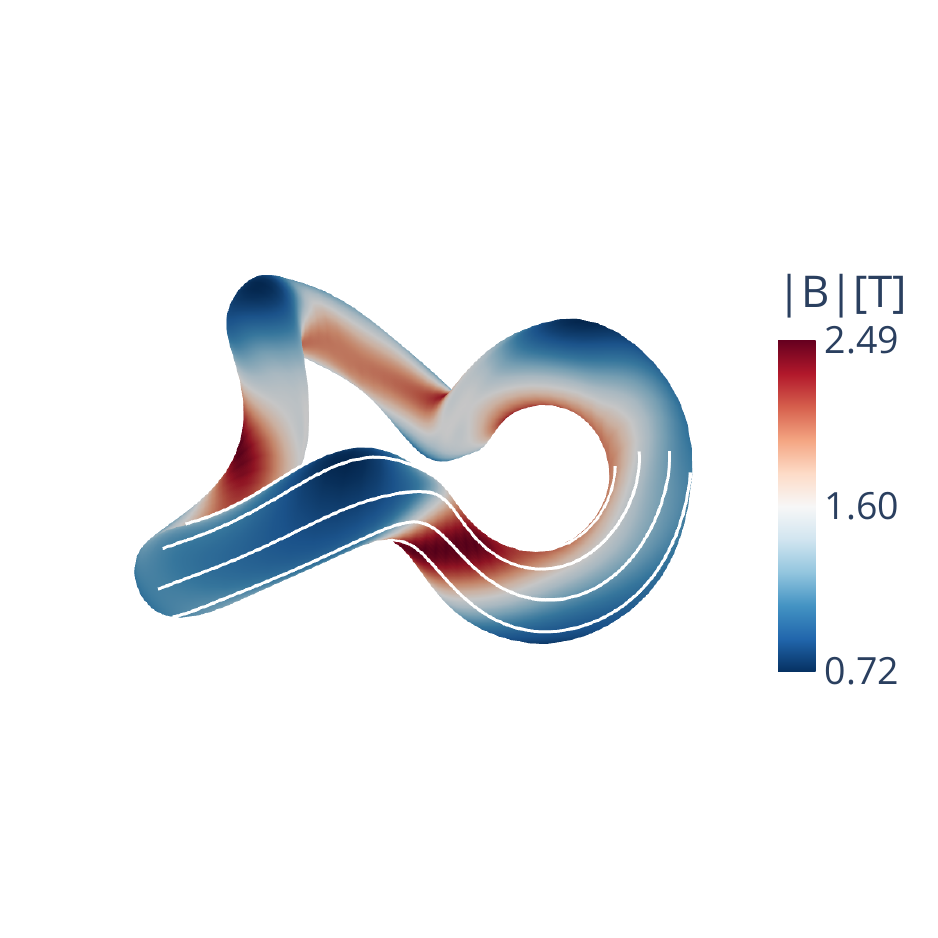}
     \includegraphics[width=0.45\textwidth]{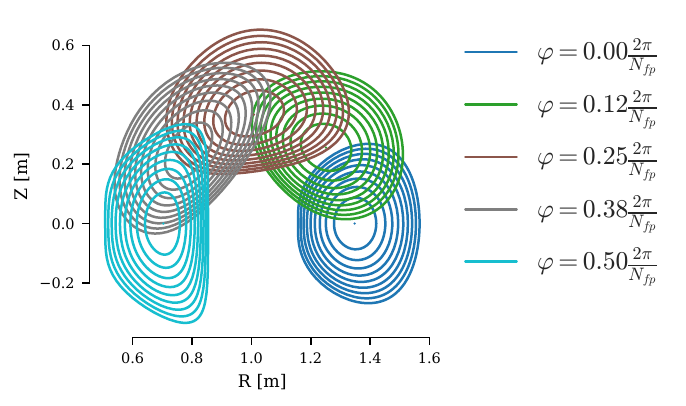}
     \caption{final geometric}
     \label{fig:image2}
   \end{subfigure}

   \begin{subfigure}[b]{0.49\textwidth}
     \centering
     \includegraphics[width=0.45\textwidth]{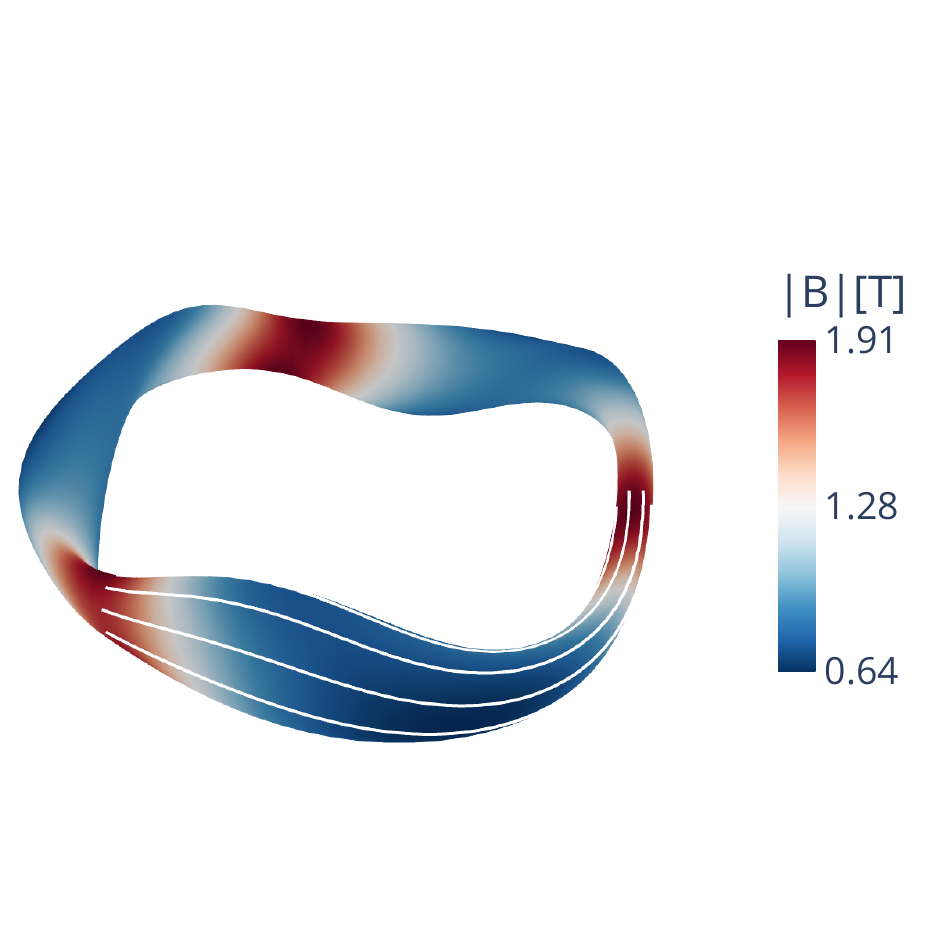}
     \includegraphics[width=0.45\textwidth]{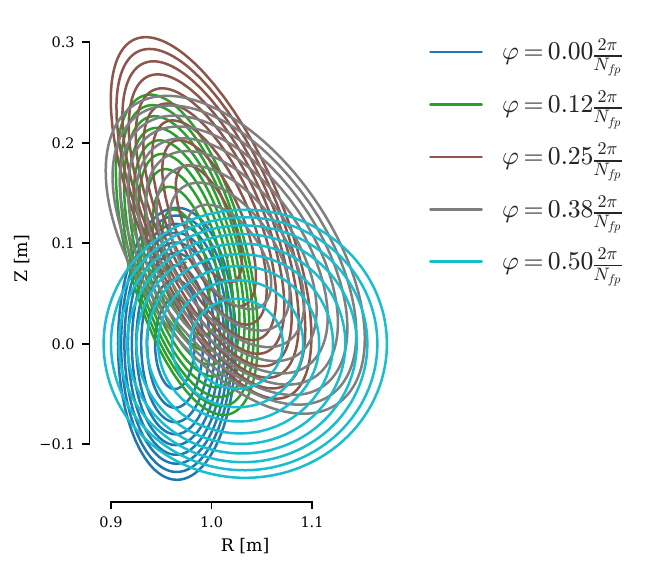}
     \caption{initial simple-to-build}
     \label{fig:image3}
   \end{subfigure}
   \hfill
   \begin{subfigure}[b]{0.49\textwidth}
     \centering
     \includegraphics[width=0.45\textwidth]{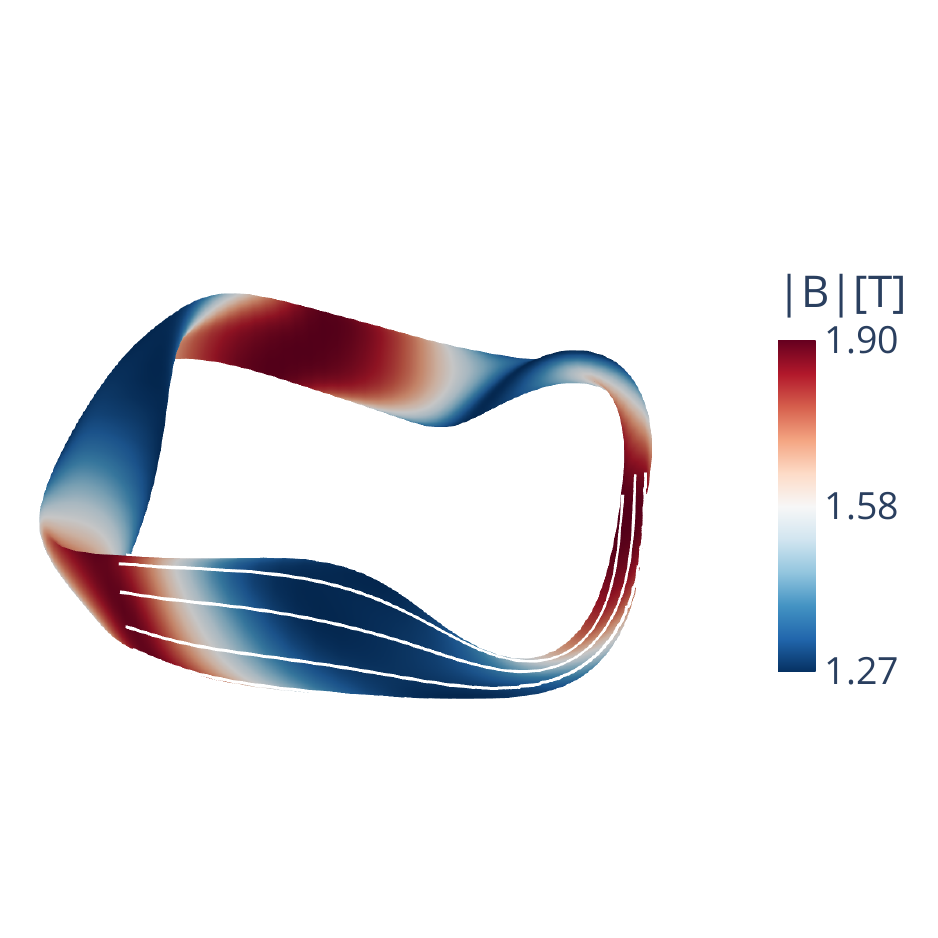}
     \includegraphics[width=0.45\textwidth]{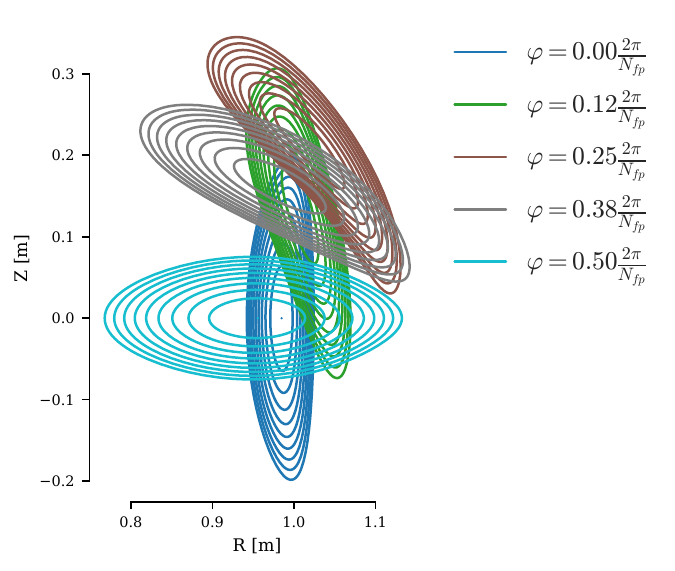}
     \caption{final simple-to-build}
     \label{fig:image4}
   \end{subfigure}

   \begin{subfigure}[b]{0.49\textwidth}
     \centering
     \includegraphics[width=0.45\textwidth]{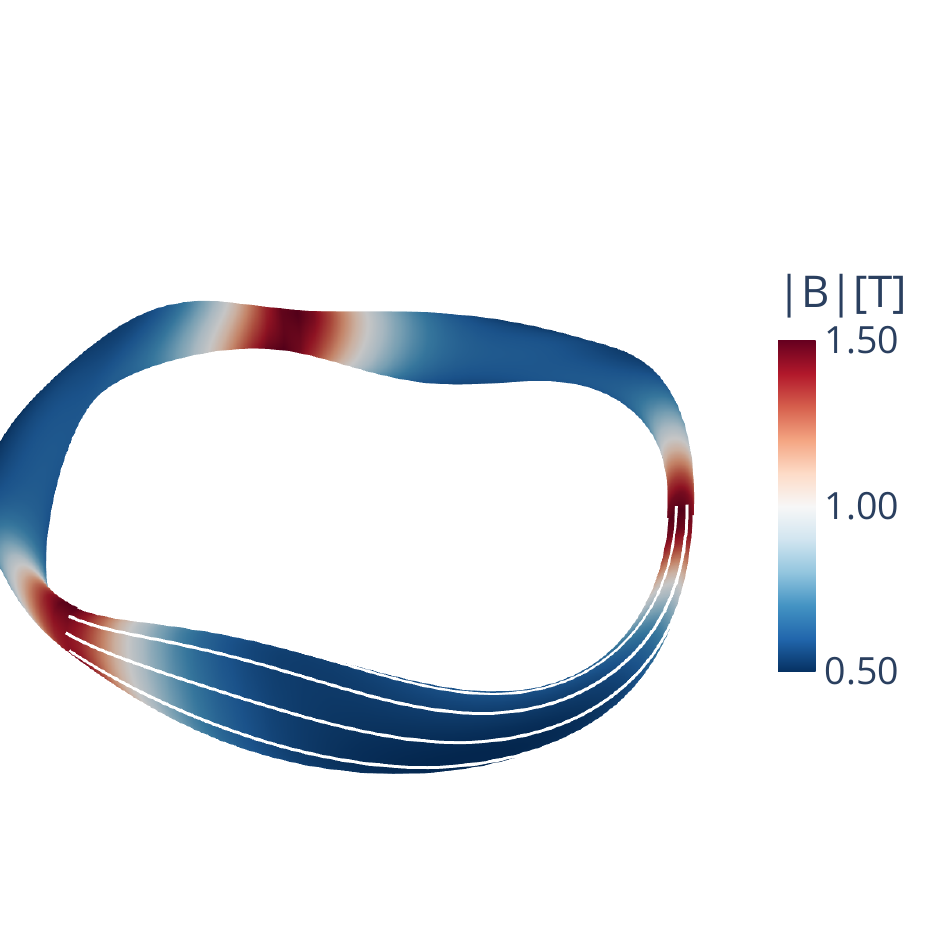}
     \includegraphics[width=0.45\textwidth]{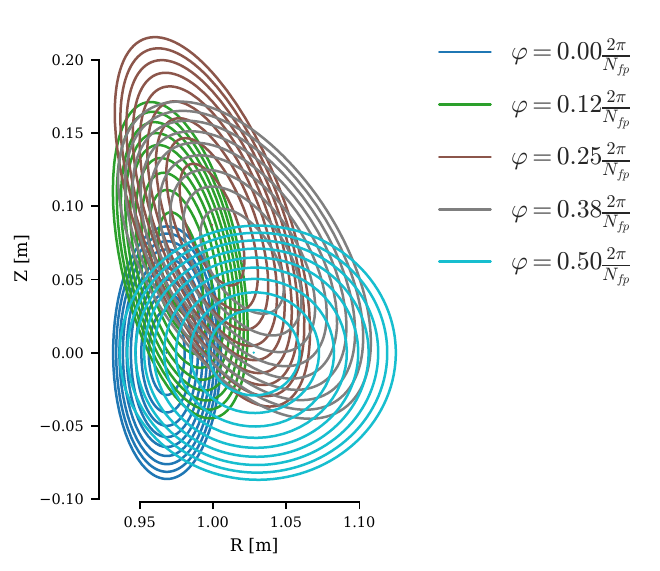}
     \caption{initial MHD-stable}
     \label{fig:image5}
   \end{subfigure}
   \hfill
   \begin{subfigure}[b]{0.49\textwidth}
     \centering
     \includegraphics[width=0.45\textwidth]{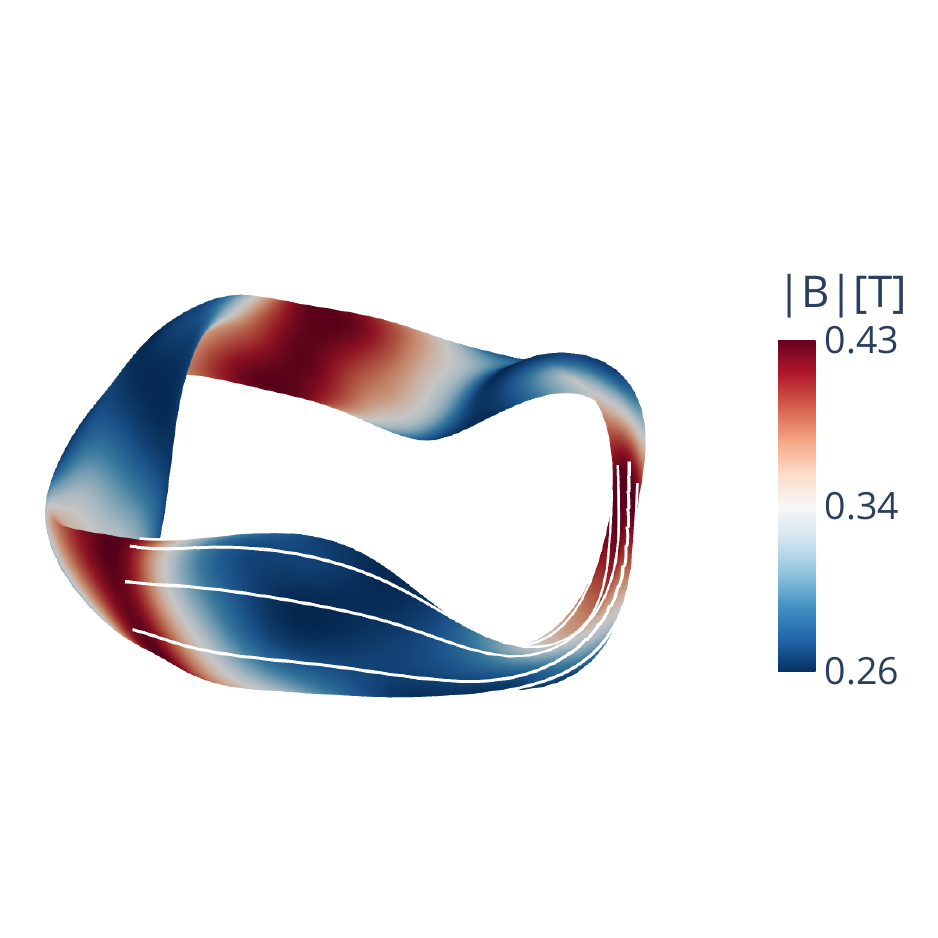}
     \includegraphics[width=0.45\textwidth]{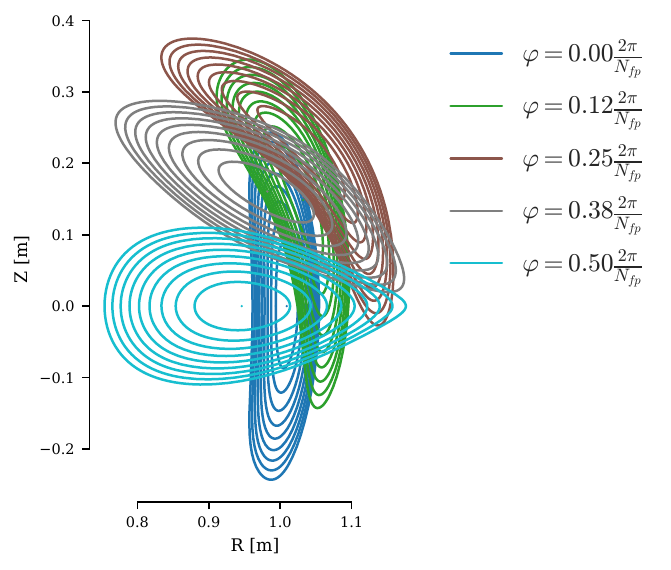}
     \caption{final MHD-stable}
     \label{fig:image6}
   \end{subfigure}
   \caption{Initial guesses and final plasma configurations optimized with ALM-NGOpt. We selected a low aspect ratio configuration from the Pareto Front of solutions for the multi-objective problem.}
   \label{fig:plasma_configurations}
\end{figure}

In \cref{fig:plasma_configurations} we show the initial and final plasma configurations for the different problems.


\subsection{Generative modeling details}\label{app:gen_model_appendix}

\begin{figure}[htbp]
  \centering
    \begin{subfigure}[b]{0.45\textwidth}
    \centering
    \includegraphics[width=\textwidth]{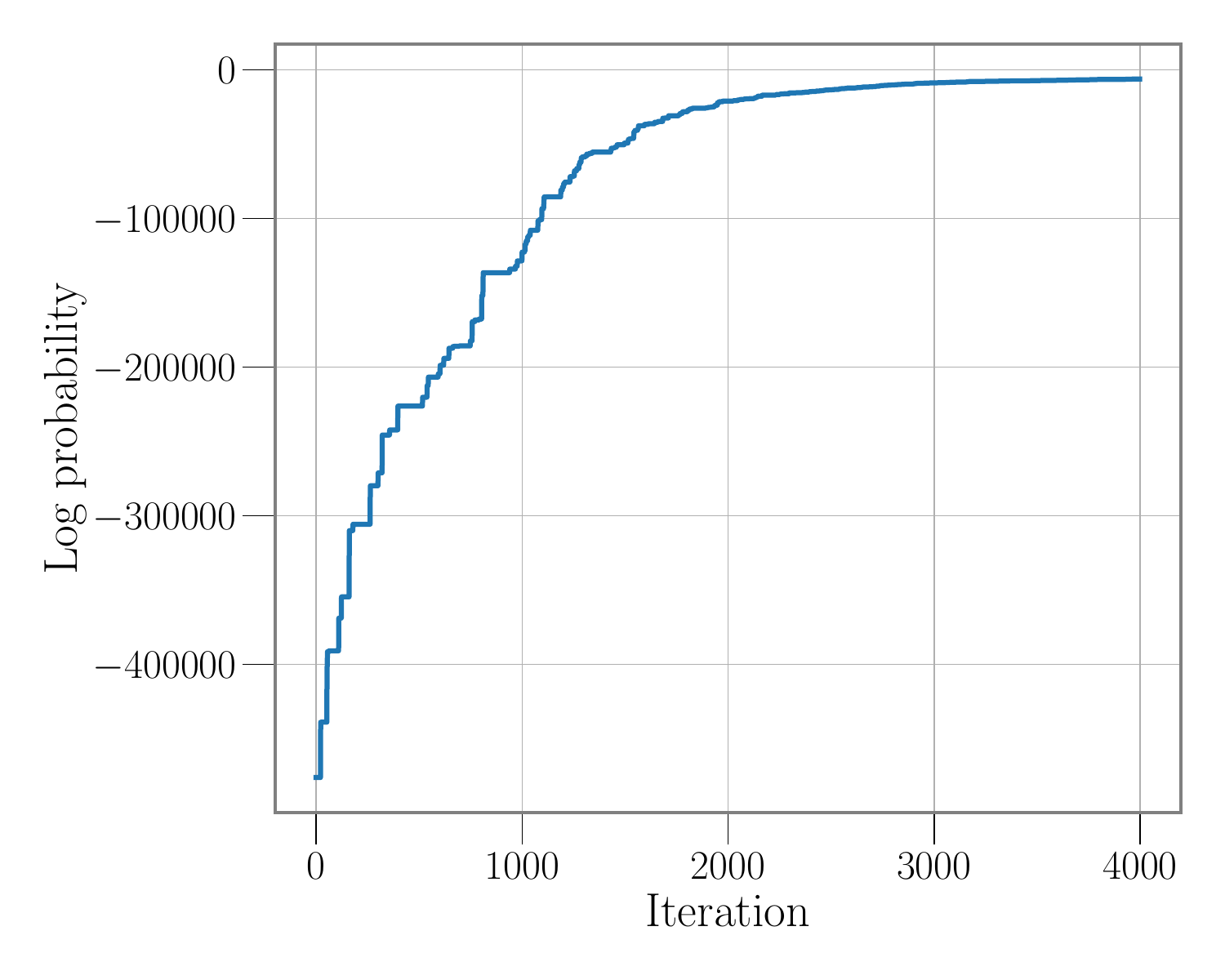}
    \caption{geometric problem.}
    \label{fig:image9}
  \end{subfigure}
  \hfill
  \begin{subfigure}[b]{0.45\textwidth}
    \centering
    \includegraphics[width=\textwidth]{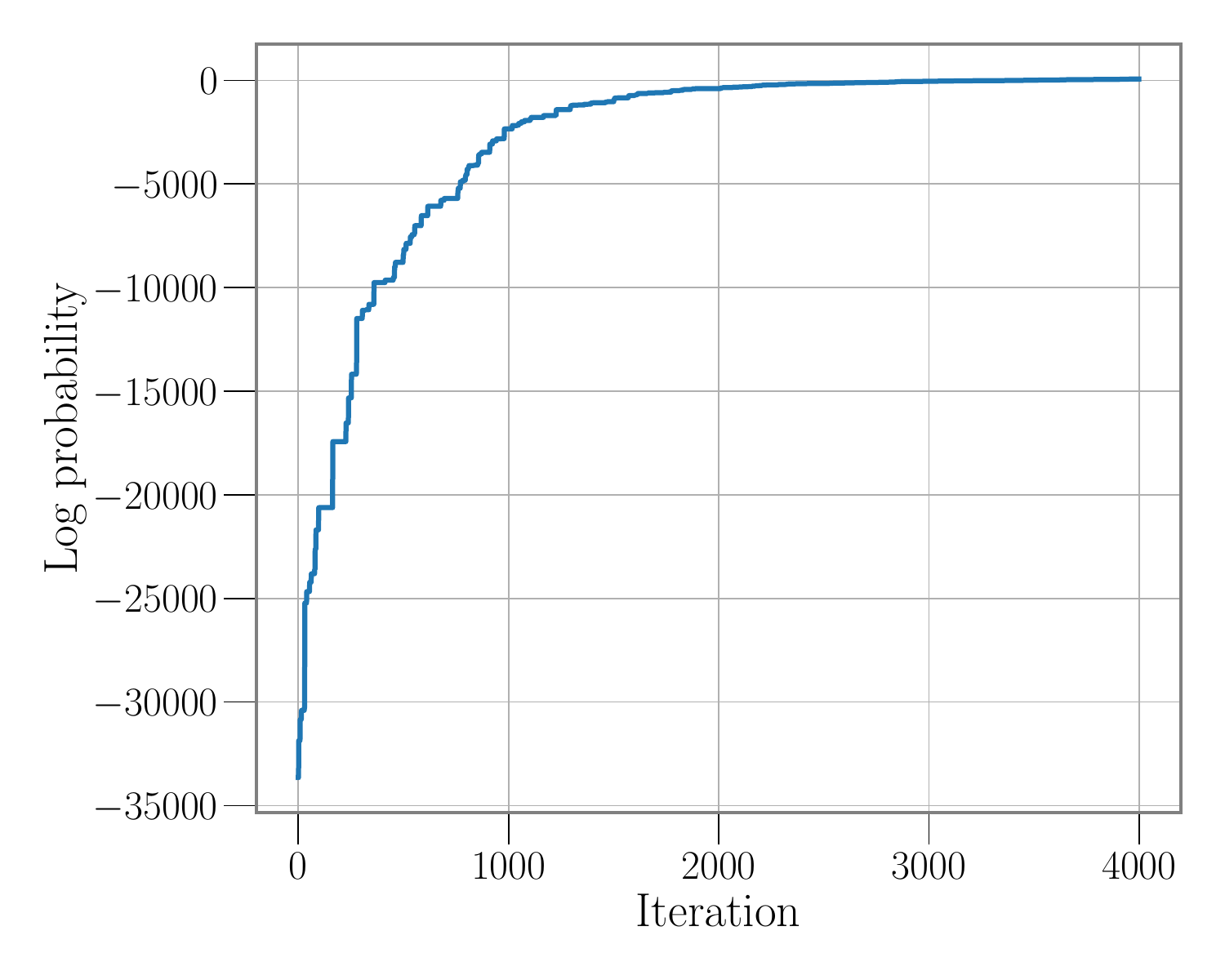}
    \caption{Simple-to-build QI problem.}
    \label{fig:image10}
  \end{subfigure}
  \caption{Trace plot of the log-posterior probability values over MCMC iterations.}
  \label{fig:mcmc_convergance}
\end{figure}

We use the Random Forest classifier and the \gls{GMM} implementations from \texttt{Scikit-learn}~\cite{scikit-learn}. We use the Random walk Metropolis-Hastings algorithm \cite{hastings1970monte} with adaptive proposal distribution \cite{haario2001adaptive} as the \gls{MCMC} sampler.
To monitor the convergence of the \gls{MCMC} sampler,
\cref{fig:mcmc_convergance} presents the log-probability of the posterior distribution evaluated at each sampled point.
The rising and stabilizing log-probability indicates convergence to high-density regions.
~\Cref{alg:feasible_generation} summarizes the formulation discussed in~\cref{sec:gen_model}.

\begin{algorithm}[t!]
\caption{Generative Inference of Feasible Configurations without Oracle Access}
\label{alg:feasible_generation}
\begin{algorithmic}[1]
\REQUIRE Dataset $\mathcal{D} = \{x_1, \ldots, x_N\} \subset \mathbb{R}^D$, constraint definition
\ENSURE Set of configurations $\{x^\ast\}$ predicted to lie in the feasible domain

// Dimensionality Reduction:
\STATE Compute PCA mapping $\Phi: \mathbb{R}^D \rightarrow \mathbb{R}^d$, where $d \ll D$
\STATE Project dataset to latent space: $Z \leftarrow \{\mathbf{z}_i = \Phi(x_i)\}_{i=1}^N$

// Feasibility Classification:
\STATE Train Random Forest classifiers $\{C_i(\mathbf{z})\}_{i=1}^{N_c}$ to predict feasibility label $y \in \{0,1\}$
\STATE Define soft-feasible region: $\tilde{\mathcal{F}} \leftarrow \bigcap_{i=1}^{N_c} \{p(C_i(\mathbf{z})=1) \geq \tau\}$, where $\tau_i = 0.8~\forall i$

// Density Estimation:
\STATE Fit Gaussian Mixture Model $\text{GMM}(\mathbf{z})$ on data restricted to $\tilde{\mathcal{F}}$

// Bayesian Refinement:
\STATE Define prior: $p(\mathbf{z}) \leftarrow \text{GMM}(\mathbf{z})$
\STATE Define quasi-likelihood: $\ell(\mathbf{z}) \leftarrow \sum_{i=1}^{N_c} \log C_i(\mathbf{z})$
\STATE Compute posterior using MCMC: $\{\mathbf{z}^\ast\} \sim p(\mathbf{z} \mid \text{feasible}) \propto \ell(\mathbf{z}) \cdot p(\mathbf{z})$
\STATE Inverse transform to original space: $x^\ast \leftarrow \Phi^{-1}(\mathbf{z}^\ast)$

// Oracle Validation:
\STATE Evaluate $x^\ast$ using \vmecpp oracle to confirm feasibility

\RETURN $\{x^\ast\}$
\end{algorithmic}
\end{algorithm}

\clearpage
\newpage
\section*{NeurIPS Paper Checklist}

\begin{enumerate}

\item {\bf Claims}
    \item[] Question: Do the main claims made in the abstract and introduction accurately reflect the paper's contributions and scope?
    \item[] Answer: \answerYes{} 
    \item[] Justification: We made claims about generating a diverse dataset of plasma configurations (see ~\cref{sec:data}). We also made claims about three optimization benchmark problems (see \cref{sec:baselines} and code), and strong baselines (see \cref{sec:baselines}). Finally we made claims about a data-driven machine learning model trained on our data for finding novel configurations (see \cref{sec:gen_model}).

\item {\bf Limitations}
    \item[] Question: Does the paper discuss the limitations of the work performed by the authors?
    \item[] Answer: \answerYes{}
    \item[] Justification: See limitations in the Discussion (\cref{sec:discussion}).
    \item[] Guidelines:
    \begin{itemize}
        \item The answer NA means that the paper has no limitation while the answer No means that the paper has limitations, but those are not discussed in the paper. 
        \item The authors are encouraged to create a separate "Limitations" section in their paper.
        \item The paper should point out any strong assumptions and how robust the results are to violations of these assumptions (\eg, independence assumptions, noiseless settings, model well-specification, asymptotic approximations only holding locally). The authors should reflect on how these assumptions might be violated in practice and what the implications would be.
        \item The authors should reflect on the scope of the claims made, e.g., if the approach was only tested on a few datasets or with a few runs. In general, empirical results often depend on implicit assumptions, which should be articulated.
        \item The authors should reflect on the factors that influence the performance of the approach. For example, a facial recognition algorithm may perform poorly when image resolution is low or images are taken in low lighting. Or a speech-to-text system might not be used reliably to provide closed captions for online lectures because it fails to handle technical jargon.
        \item The authors should discuss the computational efficiency of the proposed algorithms and how they scale with dataset size.
        \item If applicable, the authors should discuss possible limitations of their approach to address problems of privacy and fairness.
        \item While the authors might fear that complete honesty about limitations might be used by reviewers as grounds for rejection, a worse outcome might be that reviewers discover limitations that aren't acknowledged in the paper. The authors should use their best judgment and recognize that individual actions in favor of transparency play an important role in developing norms that preserve the integrity of the community. Reviewers will be specifically instructed to not penalize honesty concerning limitations.
    \end{itemize}

\item {\bf Theory assumptions and proofs}
    \item[] Question: For each theoretical result, does the paper provide the full set of assumptions and a complete (and correct) proof?
    \item[] Answer: \answerNA{} 
    \item[] Justification: We did not include any theoretical results.

    \item {\bf Experimental result reproducibility}
    \item[] Question: Does the paper fully disclose all the information needed to reproduce the main experimental results of the paper to the extent that it affects the main claims and/or conclusions of the paper (regardless of whether the code and data are provided or not)?
    \item[] Answer: \answerYes{} 
    \item[] Justification: Our code (\url{https://github.com/proximafusion/constellaration}) and the explanations found in (Sections \ref{sec:data}, \ref{sec:baselines}, \ref{sec:gen_model}) and in the Appendix, provide sufficient detail to reproduce the main claims.

\item {\bf Open access to data and code}
    \item[] Question: Does the paper provide open access to the data and code, with sufficient instructions to faithfully reproduce the main experimental results, as described in supplemental material?
    \item[] Answer: \answerYes{} 
    \item[] Justification: The code is open and available at \url{https://github.com/proximafusion/constellaration} and the dataset is also open source at \url{https://huggingface.co/datasets/proxima-fusion/constellaration}

\item {\bf Experimental setting/details}
    \item[] Question: Does the paper specify all the training and test details (e.g., data splits, hyperparameters, how they were chosen, type of optimizer, etc.) necessary to understand the results?
    \item[] Answer: \answerYes{} 
    \item[] Justification: We went in great detail to explain what optimizers we used and referenced relevant packages. See \cref{sec:baselines}, \cref{app:optimization_baselines}, \cref{sec:gen_model}, and \cref{app:gen_model_appendix}).

\item {\bf Experiment statistical significance}
    \item[] Question: Does the paper report error bars suitably and correctly defined or other appropriate information about the statistical significance of the experiments?
    \item[] Answer: \answerNA{} 
    \item[] Justification: We did not make experiments requiring error bars as we focused on 1) generating plasma boundaries, and 2) generating unique configurations that serve as baselines meeting all the constraints of our optimization problems.

\item {\bf Experiments compute resources}
    \item[] Question: For each experiment, does the paper provide sufficient information on the computer resources (type of compute workers, memory, time of execution) needed to reproduce the experiments?
    \item[] Answer: \answerYes{} 
    \item[] Justification: The computational cost of generating the dataset was outline on a per-optimization method approach in \cref{sec:data}. For the optimization baseline, we provided the compute resources (number of vCPUs and time of execution) in the captions of Tables~\ref{tab:comparison_optimization_methods} and~\ref{tab:multi_objective}.
    
\item {\bf Code of ethics}
    \item[] Question: Does the research conducted in the paper conform, in every respect, with the NeurIPS Code of Ethics \url{https://neurips.cc/public/EthicsGuidelines}?
    \item[] Answer: \answerYes{} 
    \item[] Justification: We reviewed the NeurIPS Code of Ethics and believe that we adhere to it.

\item {\bf Broader impacts}
    \item[] Question: Does the paper discuss both potential positive societal impacts and negative societal impacts of the work performed?
    \item[] Answer: \answerYes{} 
    \item[] Justification: In the Abstract and Introduction we reflect on the potential of fusion energy and the impact that stellarator optimization can have for a future of clean energy.
    
\item {\bf Safeguards}
    \item[] Question: Does the paper describe safeguards that have been put in place for responsible release of data or models that have a high risk for misuse (e.g., pretrained language models, image generators, or scraped datasets)?
    \item[] Answer: \answerNA{} 
    \item[] Justification: We don't believe our code and dataset of plasma boundaries and ideal \gls{MHD} equilibria caries any risk.

\item {\bf Licenses for existing assets}
    \item[] Question: Are the creators or original owners of assets (e.g., code, data, models), used in the paper, properly credited and are the license and terms of use explicitly mentioned and properly respected?
    \item[] Answer: \answerYes{} 
    \item[] Justification: Both the code and the dataset we release use the MIT license as specified in their corresponding repositories. The list of dependencies also have very permissive licenses.

\item {\bf New assets}
    \item[] Question: Are new assets introduced in the paper well documented and is the documentation provided alongside the assets?
    \item[] Answer: \answerYes{} 
    \item[] Justification: The code and the dataset have thorough README files. Moreover, the paper describes the data generation process in detail (see \cref{sec:data}).

\item {\bf Crowdsourcing and research with human subjects}
    \item[] Question: For crowdsourcing experiments and research with human subjects, does the paper include the full text of instructions given to participants and screenshots, if applicable, as well as details about compensation (if any)? 
    \item[] Answer: \answerNA{} 
    \item[] Justification: No crowdsorcing involved.

\item {\bf Institutional review board (IRB) approvals or equivalent for research with human subjects}
    \item[] Question: Does the paper describe potential risks incurred by study participants, whether such risks were disclosed to the subjects, and whether Institutional Review Board (IRB) approvals (or an equivalent approval/review based on the requirements of your country or institution) were obtained?
    \item[] Answer: \answerNA{} 
    \item[] Justification: Did not involve any crowdsourcing.

\item {\bf Declaration of LLM usage}
    \item[] Question: Does the paper describe the usage of LLMs if it is an important, original, or non-standard component of the core methods in this research? Note that if the LLM is used only for writing, editing, or formatting purposes and does not impact the core methodology, scientific rigorousness, or originality of the research, declaration is not required.
    \item[] Answer: \answerNA{}.
    \item[] Justification: Only used LLMs for text editing.
\end{enumerate}

\end{document}